\documentclass[journal]{IEEEtran}
\usepackage{times}
\usepackage{epsfig}
\usepackage{graphicx}
\usepackage{amsmath}
\usepackage{amssymb}

\usepackage{textcomp,gensymb}
\usepackage{url}            
\usepackage{booktabs}       
\usepackage{amsfonts}       
\usepackage{nicefrac}       
\usepackage{color,colortbl}
\usepackage{pifont}
\newcommand{\cmark}{\ding{51}}%
\usepackage{cite}

\usepackage{silence}
\usepackage{multirow}
\usepackage{bm}

\usepackage{makecell}
\usepackage{boldline}
\setcellgapes{3pt}
\usepackage{arydshln}
\usepackage{tabu}
\usepackage{subfigure}
\usepackage{balance}

\usepackage[pagebackref=true,breaklinks=true,colorlinks,bookmarks=false]{hyperref}
\usepackage{cleveref}
\crefformat{section}{\S#2#1#3} 
\crefformat{subsection}{\S#2#1#3}
\crefformat{subsubsection}{\S#2#1#3}

\begin{document}

\title{Noisy-As-Clean: Learning Self-supervised Denoising from the Corrupted Image}

\author{Jun Xu,
        Yuan Huang,
        Ming-Ming Cheng,~\IEEEmembership{Senior Member,~IEEE,}
        Li Liu,
        Fan Zhu,
        Zhou Xu,
        Ling Shao
\thanks{
This work was supported in part by the Major Project for New Generation of AI under Grant No. 2018AAA0100400, NSFC (61922046), 
and Tianjin Natural Science Foundation (18ZXZNGX00110). 
($^*$Corresponding author: M.-M. Cheng)
}
\thanks{
J. Xu and M.-M. Cheng are with TKLNDST, College of Computer Science, Nankai University, Tianjin, China (E-mail: csjunxu@nankai.edu.cn, cmm@nankai.edu.cn).
Y. Huang is with School of Electronic and Information Engineering, Xi'an Jiaotong University, Xi'an, China.
L. Liu, F. Zhu, and L. Shao are with Inception Institute of Artificial Intelligence (IIAI) and Mohamed bin Zayed University of Artificial Intelligence (MBZUAI), Abu Dhabi, UAE.
Zhou Xu is with School of Big Data and Software Engineering, Chongqing University, Chongqing, China.
The first two authors contribute equally. 
}
}

\IEEEtitleabstractindextext{%
\begin{abstract}
Supervised deep networks have achieved promising performance on image denoising, by learning image priors and noise statistics on plenty pairs of noisy and clean images.
Unsupervised denoising networks are trained with only noisy images.
However, for an unseen corrupted image, both supervised and unsupervised networks ignore either its particular image prior, the noise statistics, or both.
That is, the networks learned from external images inherently suffer from a domain gap problem: the image priors and noise statistics are very different between the training and test images.
This problem becomes more clear when dealing with the signal dependent realistic noise.
To circumvent this problem, in this work, we propose a novel ``Noisy-As-Clean'' (NAC) strategy of training self-supervised denoising networks.
Specifically, the corrupted test image is directly taken as the ``clean'' target, while the inputs are synthetic images consisted of this corrupted image and a second and similar corruption.
A simple but useful observation on our NAC is: \textsl{as long as the noise is weak, it is feasible to learn an self-supervised network only with the corrupted image, approximating the optimal parameters of a supervised network learned with pairs of noisy and clean images}.
Experiments on synthetic and realistic noise removal demonstrate that, the DnCNN and ResNet networks trained with our self-supervised NAC strategy achieve comparable or better performance than the original ones and previous supervised/unsupervised/self-supervised networks.
The code is publicly available at \url{https://github.com/csjunxu/Noisy-As-Clean}.
\end{abstract}

}

\maketitle

\IEEEdisplaynontitleabstractindextext

\IEEEpeerreviewmaketitle

\section{Introduction}
\label{sec:introduction}

\IEEEPARstart{I}{mage} denoising is an ill-posed inverse problem to recover a \textsl{clean} image $\mathbf{x}$ from the \textsl{observed} noisy image $\mathbf{y}=\mathbf{x}+\mathbf{n}_{o}$, where $\mathbf{n}_{o}$ is the \textsl{observed} corrupted noise.\ 
One popular assumption on $\mathbf{n}$ is the additive white Gaussian noise (AWGN) with standard deviation (std) $\sigma$, which serves as a perfect test bed for supervised networks in the deep learning era~\cite{vggnet,googlenet,resnet}.\
Supervised networks~\cite{nlnet,dncnn,n3net} learn the image priors and noise statistics on plenty pairs of clean and corrupted images, and achieve promising denoising performance on the images with similar priors and noise statistics (e.g., AWGN).

\begin{figure}[t]
\centering
\subfigure{
\begin{minipage}{0.23\textwidth}
\includegraphics[width=1\textwidth]{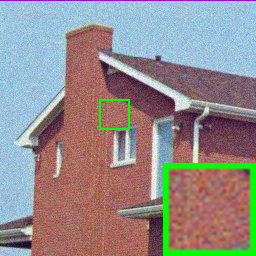}\vspace{-1mm}
\centering{(a) Noisy: 24.62dB/0.4595}
\end{minipage}
\begin{minipage}{0.23\textwidth}
\includegraphics[width=1\textwidth]{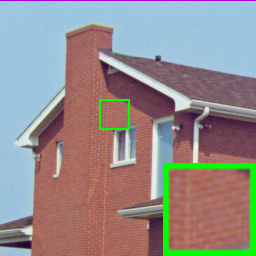}\vspace{-1mm}
\centering{(b) Clean Image}
\end{minipage}
}\vspace{-2.5mm}
\subfigure{
\begin{minipage}{0.23\textwidth}
\includegraphics[width=1\textwidth]{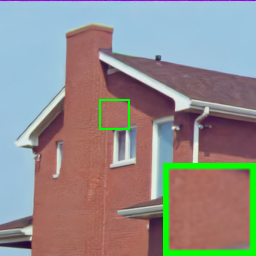}\vspace{-1mm}
\centering{(c) DnCNN~\cite{dncnn}: 34.23dB/0.8695}
\end{minipage}
\begin{minipage}{0.23\textwidth}
\includegraphics[width=1\textwidth]{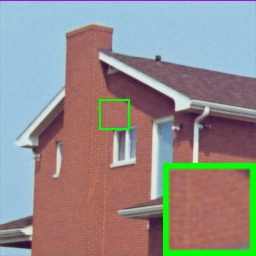}\vspace{-1mm}
\centering{(d) DnCNN+NAC: \textbf{35.80}dB/\textbf{0.9116}}
\end{minipage}
}
\vspace{-2mm}
\caption{Denoised images and PSNR/SSIM results of DnCNN~\cite{dncnn} (c) and DnCNN trained by our NAC strategy (``DnCNN+NAC'') (d) on the color image \textsl{House} (b) corrupted by AWGN noise ($\sigma=15$) (a).}
\vspace{-2.5mm}
\label{f-example}
\end{figure}

With advances on AWGN noise removal~\cite{mlp,dncnn,n3net}, a natural question arises is how these denoising networks can exert their effect on real noisy photographs.\
Realistic noise is signal dependent and more complex than AWGN~\cite{crosschannel2016,dnd2017,PolyUdataset}.\ 
Thus, previous supervised denoising networks unavoidably suffer from a \textsl{domain gap problem}: both the image priors and noise statistics in training are different from those of the real-world test images.\ 
Recently, several unsupervised~\cite{gcbd,noise2noise,dip,noise2void} and self-supervised~\cite{noise2self,ss2019} networks have been developed to get rid of the dependence on clean images, which are difficult to be obtained in real-world scenarios.\
However, unsupervised networks are subjected to the gap on either image priors or noise statistics, while self-supervised suffer from the gap on noise statistics, between the external images for training and the corrupted ones for test.\ 
Besides, several networks~\cite{noise2noise,dip} succeed on the zero-mean noise.\
But the realistic noise in real-world images is not necessarily zero-mean~\cite{crosschannel2016,dnd2017,sidd2018}.

To alleviate the domain gap on image priors and noise statistics between training and test images, in this paper, we propose a ``Noisy-As-Clean'' (NAC) strategy for training self-supervised denoising networks.\ 
In our NAC, we directly train an image-specific network by taking the corrupted image $\mathbf{y}=\mathbf{x}+\mathbf{n}_{o}$ as the ``\textsl{clean}'' target.\ 
Thus, the domain gap on image priors are largely bridged by our NAC.\ 
To reduce the gap on noise statistics, for the target corrupted image $\mathbf{y}$, we take as the input of our NAC a \textsl{simulated} noisy image $\mathbf{z}=\mathbf{y}+\mathbf{n}_{s}$ consisting of the corrupted image $\mathbf{y}$ and a \textsl{simulated} noise $\mathbf{n}_{s}$, which is statistically close to the corrupted noise $\mathbf{n}_{o}$ in $\mathbf{y}$.\
By this way, our NAC network learns to clean up the \textsl{simulated} noise $\mathbf{n}_{s}$ from the doubly corrupted image $\mathbf{z}$ during training, and thus is able to remove the noise $\mathbf{n}_{o}$ from the corrupted image $\mathbf{y}$ during test.

A simple but useful observation about our NAC strategy is: \textsl{as long as the corrupted noise is ``weak'', it is feasible to train a self-supervised denoising network only with the corrupted test image, and the learned parameters are very close to those of a \textsl{supervised} network trained with a pair of the corrupted image and its clean version}.\ Though being very simple, our NAC strategy is very effective for image denoising.\ In Figure~\ref{f-example}, we compare the denoised images by the vanilla DnCNN~\cite{dncnn} and the DnCNN trained with our NAC (DnCNN+NAC), on the image ``House'' corrupted by AWGN ($\sigma=15$).\
We observe that the ``DnCNN+NAC'' achieves better visual quality and higher PSNR/SSIM results than DnCNN~\cite{dncnn}, which is trained on plenty of noisy and clean image pairs.\ %
Experiments on diverse benchmarks demonstrate that, when trained with our NAC strategy, the DnCNN~\cite{dncnn} and ResNet~\cite{resnet} in Deep Image Prior (DIP)~\cite{dip} achieve comparable or better performance than supervised denoising networks on synthetic and real-world noisy images.\ 
Our work reveals that, \textsl{when the noise is ``weak''}, a self-supervised network trained directly on the corrupted image can obtain comparable or even better performance than supervised networks on image denoising.

In summary, our contribution are mainly three-fold:
\begin{itemize}
    \item We propose a ``Noisy-As-Clean'' (NAC) strategy for training self-supervised denoising networks.
    \item We provide a theoretical background of our NAC strategy, and implement the DnCNN~\cite{dncnn} and ResNet in DIP~\cite{dip} into self-supervised networks by our NAC for effective image denoising.
    \item Experiments on synthetic and real-world benchmarks show that, on weak noise, the DnCNN and ResNet in~\cite{dip} trained by our NAC achieve comparable or even better performance than the comparison denoising networks.
\end{itemize}

The remaining parts of this paper are organized as follows.
In \S\ref{sec:related}, we introduce the related work.
In \S\ref{sec:nas}, we present the theoretical background of our NAC strategy for self-supervised image denoising.\
In \S\ref{sec:self-sup}, we implement the DnCNN~\cite{dncnn} and ResNet used in~\cite{dip} as self-supervised networks by our NAC.\
Extensive experiments are conducted in \S\ref{sec:exp} demonstrate that, the DnCNN and ResNet networks trained by our NAC achieve comparable or even better performance than previous supervised image denoising networks on benchmark synthetic and real-world datasets.\
Conclusion is given in \S\ref{sec:con}.



\section{Related Work}
\label{sec:related}

In Table~\ref{t1}, we summarize several state-of-the-art supervised~\cite{dncnn,cbdnet}, unsupervised~\cite{noise2noise,gcbd,noise2void} and self-supervised~\cite{dip,noise2self,ss2019} networks, image priors, and noise statistics.\ 
In this work, to bridge the \textsl{domain gap problem}, we propose a ``Noisy-As-Clean'' strategy to learn the image-specific internal prior and noise statistics directly from the corrupted test image.\ 

\noindent
\textbf{Supervised denoising networks} are trained with plenty pairs of noisy and clean images.\
This category of networks can learn external image priors and noise statistics from the training data.\ 
Several methods~\cite{dncnn,n3net,nlrn2018} have been developed with achieving promising performance on AWGN noise removal, where the statistics of training and test noise are similar.\ 
However, due to the aforementioned \textsl{domain gap problem}, the performance of these networks degrade severely on real-world noisy images~\cite{crosschannel2016,dnd2017,PolyUdataset}.\

\noindent
\textbf{Unsupervised and self-supervised denoising networks} are developed to remove the need on plenty of clean images.\ Along this direction, Noise2Noise (N2N)~\cite{noise2noise} trains the network between pairs of corrupted images with the same scene, but independently sampled noise.\ This work is feasible to learn external image priors and noise statistics from the training data.\ However, in real-world scenarios, it is difficult to collect large amounts of paired images with independent corruption for training.\ Noise2Void (N2V)~\cite{noise2void} predicts a pixel from its surroundings by learning blind-spot networks, but it still suffers from the domain gap on image priors between the training images and test images.\ This work assumes that the corruption is zero-mean and independent between pixels.\ However, as mentioned in Noise2Self (N2S)~\cite{noise2self}, N2V~\cite{noise2void} significantly degrades the training efficiency and denoising performance at test time.\ Recently, Deep Image Prior (DIP)~\cite{dip} reveals that the network structure can resonate with the natural image priors, and can be utilized in image restoration without external images.\ However, it is not practical to select a suitable network and early-stop its training at right moments for each corrupted image.
Self-supervised denoisers~\cite{noise2self,ss2019} employ explicit corruption models, and train the networks only with the corrupted image itself.\ 
In this work, we utilize the helpful noise model to learn self-supervised denoising networks for real-world image denoising.\

\begin{table}[t]
\centering
\caption{\textbf{Summary of representative networks for image denoising}.
\textbf{S.}: Supervised networks.
\textbf{U.}: Unsupervised networks.
\textbf{SS.}: Self-supervised networks.
\textbf{Pub.}: Publication.
\textbf{Int.}: Internal image priors.
\textbf{Ext.}: External image priors.
\textbf{Stat.}: Statistics.
The networks with ``\cmark'' are able to learn the noise statistics from training data.
}
\begin{tabular}{c|rl|c|c}
\Xhline{1pt}
\rowcolor[rgb]{ .85,  .9,  0.95}
& 
&
& Image
& Noise
\\
\rowcolor[rgb]{ .85,  .9,  0.95}
\multicolumn{1}{c|}{\multirow{-2}{*}{Type}}
& \multicolumn{1}{c}{\multirow{-2}{*}{Method}}
& \multicolumn{1}{l|}{\multirow{-2}{*}{Year'Pub.}} 
& \multicolumn{1}{c|}{Prior}
& Stat.
\\
\hline
\hline
\multicolumn{1}{c|}{\multirow{2}{*}{S.}} & DnCNN~\cite{dncnn}
& 17'TIP & Ext. & \cmark
\\
& CBDNet~\cite{cbdnet}
& 19'CVPR & Ext. & \cmark
\\
\hline
\hline
\multicolumn{1}{c|}{\multirow{3}{*}{U.}}& Noise2Noise~\cite{noise2noise} & 18'ICML & Ext. & \cmark
\\
& GAN-CNN~\cite{gcbd} & 18'CVPR & Ext. & \cmark
\\
& Noise2Void~\cite{noise2void} & 19'CVPR & Ext. &
\\
\hline
\hline
\multicolumn{1}{c|}{\multirow{4}{*}{SS.}}
&
Deep Image Prior~\cite{dip} & 18'CVPR & Int. &
\\
& Noise2Self\ \ ~\cite{noise2self} & 19'ICML & Ext. &
\\
& Self-Supervised~\cite{ss2019} & 19'NeurIPS & Ext. &
\\
& 
Noisy-As-Clean (Ours) & 20'Submit & Int. & \cmark
\\
\hline
\end{tabular}%
\vspace{-2.5mm}
\label{t1}%
\end{table}%
\begin{figure*}[htp]
\vspace{-0mm}
\centering
\raisebox{-0.15cm}{\includegraphics[width=1\textwidth]{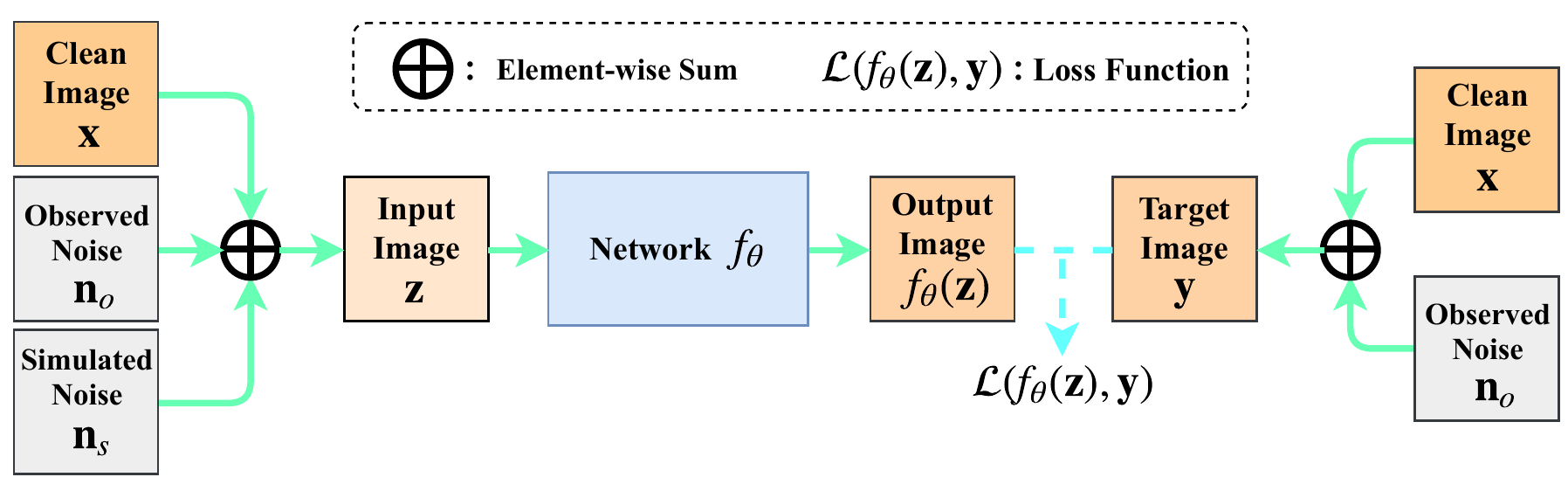}}
\vspace{-2.5mm}
\caption{\textbf{Proposed ``Noisy-As-Clean'' strategy for training self-supervised image denoising networks}.\ 
In our NAC strategy, we take the \textsl{observed} noisy image $\mathbf{y}=\mathbf{x}+\mathbf{n}_{o}$ as the ``clean'' target, and take the \textsl{simulated} noisy image $\mathbf{z}=\mathbf{y}+\mathbf{n}_{s}$ as the input.
We do not regard the clean image $\mathbf{x}$ as target.
After training, the inference is performed on the target noisy image $\mathbf{y}=\mathbf{x}+\mathbf{n}_{o}$.
}
\vspace{-2.5mm}
\label{f1}
\end{figure*}

\noindent
\textbf{Internal and external image priors} are widely used for diverse image restoration tasks~\cite{mcwnnm,foe,pgpd}.\ Internal priors are directly learned from the input test image itself, such as the multi-scale priors~\cite{ksvd,STAR2020,cvid2020}, image-specific details~\cite{iraniinternal,Liang_2018_CVPR}, and non-local self similarity~\cite{mcwnnm,twsc,NLH2020}.\ 
The external ones are learned on external natural images~\cite{epll,pgpd,gid2018}.\ 
Internal priors are adaptive to its image contents, but somewhat affected by the corruptions~\cite{ksvd,iraniinternal}.\ 
By contrast, the external priors are effective for restoring images with general contents, but may not be optimal for specific test image~\cite{epll,pgpd,chen2017trainable}.\

\noindent
\textbf{Noise statistics} is of key importance for image denoising.\ The AWGN noise is one typical noise with widespread study.\ Recently, researchers shift more attention to the realistic noise produced in camera sensors~\cite{dnd2017,sidd2018}, which is usually modeled as mixed Poisson and Gaussian distribution~\cite{poissongaussian}.\ The Poisson component mainly comes from the irregular photons hitting the sensor~\cite{liu2006noise}, while Gaussian noise is majorly produced by dark current~\cite{crosschannel2016}.\ Though performing well on the synthetic noise being trained with, supervised denoisers~\cite{dncnn,nlrn2018,cbdnet} still suffer from the \textsl{domain gap problem} when processing the real-world noisy images.

\section{Theoretical Background of ``Noisy-As-Clean''}
\label{sec:nas}
Training a supervised network $f_{\theta}$ (parameterized by $\theta$) requires many pairs $\{(\mathbf{y}_{i},\mathbf{x}_{i})\}$ of noisy image $\mathbf{y}_{i}$ and clean image $\mathbf{x}_{i}$,
by minimizing an empirical loss function $\mathcal{L}$ as
\begin{equation}
\arg\min_{\theta}\sum_{i=1}\mathcal{L}(f_{\theta}(\mathbf{y}_{i}),\mathbf{x}_{i}).
\end{equation}
Assuming that the 
probability of occurrence for pair $(\mathbf{y}_{i},\mathbf{x}_{i})$ is $p(\mathbf{y}_{i},\mathbf{x}_{i})$, then statistically we have 
\begin{equation}
\label{eqn:e2}
\begin{split}
\theta^{*}
&=\arg\min_{\theta}\sum_{i=1}p(\mathbf{y}_{i},\mathbf{x}_{i})\mathcal{L}(f_{\theta}(\mathbf{y}_{i}),\mathbf{x}_{i})
\\
&=
\arg\min_{\theta}\mathbb{E}_{(\mathbf{y},\mathbf{x})}[\mathcal{L}(f_{\theta}(\mathbf{y}),\mathbf{x})],
\end{split}
\end{equation}
where $\mathbf{y}$ and $\mathbf{x}$ are random variables of noisy and clean images, respectively.\ 
The paired variables $(\mathbf{y},\mathbf{x})$ are dependent, and their relationship is $\mathbf{y}=\mathbf{x}+\mathbf{n}_{o}$, where $\mathbf{n}_{o}$ is the random variable of \textsl{observed} noise.\ 
By exploring the dependence of
$p(\mathbf{y}_{i},\mathbf{x}_{i})=p(\mathbf{x}_{i})p(\mathbf{y}_{i}|\mathbf{x}_{i})$,
Eqn.~(2) is equivalent to
\begin{equation}
\label{e3}
\begin{split}
\theta^{*}
&=\arg\min_{\theta}\sum_{i=1}p(\mathbf{x}_{i})p(\mathbf{y}_{i}|\mathbf{x}_{i})\mathcal{L}(f_{\theta}(\mathbf{y}_{i}),\mathbf{x}_{i})
\\
&=
\arg\min_{\theta}\mathbb{E}_{\mathbf{x}}[\mathbb{E}_{\mathbf{y}|\mathbf{x}}[\mathcal{L}(f_{\theta}(\mathbf{y}),\mathbf{x})]]
.
\end{split}
\end{equation}
This indicates that the network $f_{\theta}$ can minimize the loss function by solving Eqn.~(3) separately for each clean image.\ 

Different with the ``zero-mean'' assumption in~\cite{noise2noise,noise2void}, here we study a more practical assumption on noise statistics, i.e., \textsl{the expectation $\mathbb{E}[\mathbf{x}]$ and variance} $\text{Var}[\mathbf{x}]$ \textsl{of signal intensity are much stronger than those of noise $\mathbb{E}[\mathbf{n}_{o}]$ and} $\text{Var}[\mathbf{n}_{o}]$ (negligible but not necessarily zero):
\begin{equation}
\label{e4}
\mathbb{E}[\mathbf{x}]
\gg
\mathbb{E}[\mathbf{n}_{o}]
,
\ 
\text{Var}[\mathbf{x}]
\gg
\text{Var}[\mathbf{n}_{o}]
.
\end{equation}
This is actually valid in real-world scenarios, since we can clearly observe the contents in most real photographs, \textsl{with little influence of the noise}.\ 
The noise therein is often modeled by zero-mean Gaussian or mixed Poisson and Gaussian (for realistic noise).\ 
Hence, the noisy image $\mathbf{y}$ should have similar expectation with the clean image $\mathbf{x}$:
\begin{equation}
\label{e5}
\mathbb{E}[\mathbf{y}]
=
\mathbb{E}[\mathbf{x}+\mathbf{n}_{o}]
=
\mathbb{E}[\mathbf{x}]+\mathbb{E}[\mathbf{n}_{o}]
\approx
\mathbb{E}[\mathbf{x}].
\end{equation}
Now we add \textsl{simulated} noise $\mathbf{n}_{s}$ to the \textsl{observed} noisy image $\mathbf{y}$, and generate a new noisy image $\mathbf{z}=\mathbf{y}+\mathbf{n}_{s}$.\ 
We assume that $\mathbf{n}_{s}$ is statisticly close to $\mathbf{n}_{o}$, i.e., $\mathbb{E}[\mathbf{n}_{s}]\approx\mathbb{E}[\mathbf{n}_{o}]$ and $\text{Var}[\mathbf{n}_{s}]\approx\text{Var}[\mathbf{n}_{o}]$.\ 
Then we have
\begin{equation}
\label{e6}
\mathbb{E}[\mathbf{z}]\gg\mathbb{E}[\mathbf{n}_{s}],
\ 
\text{Var}[\mathbf{z}]\gg\text{Var}[\mathbf{n}_{s}]
.
\end{equation}
Therefore, the \textsl{simulated} noisy image $\mathbf{z}$ has similar expectation with the \textsl{observed} noisy image $\mathbf{y}$:
\begin{equation}
\label{e7}
\mathbb{E}[\mathbf{z}]
=
\mathbb{E}[\mathbf{y}+\mathbf{n}_{s}]
\approx
\mathbb{E}[\mathbf{y}]
.
\end{equation}
By the \textsl{Law of Total Expectation}~\cite{billingsley1995probability}, we have
\begin{equation}
\label{e8}
\mathbb{E}_{\mathbf{y}}[\mathbb{E}_{\mathbf{z}}[\mathbf{z}|\mathbf{y}]]
=
\mathbb{E}[\mathbf{z}]
\approx
\mathbb{E}[\mathbf{y}]
=
\mathbb{E}_{\mathbf{x}}[\mathbb{E}_{\mathbf{y}}[\mathbf{y}|\mathbf{x}]]
.
\end{equation} 
\begin{figure*}[htbp]
\centering
\subfigure{
\begin{minipage}[t]{0.24\textwidth}
\centering
\raisebox{-0.15cm}{\includegraphics[width=1\textwidth]{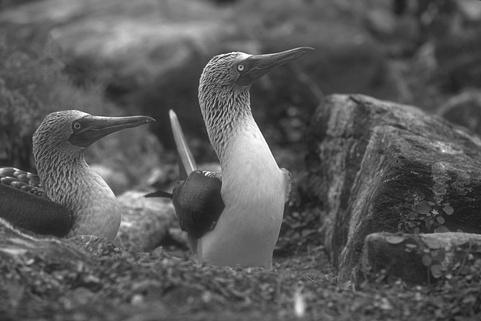}}
{\footnotesize (a) Clean Image}
\end{minipage}
\begin{minipage}[t]{0.24\textwidth}
\centering
\raisebox{-0.15cm}{\includegraphics[width=1\textwidth]{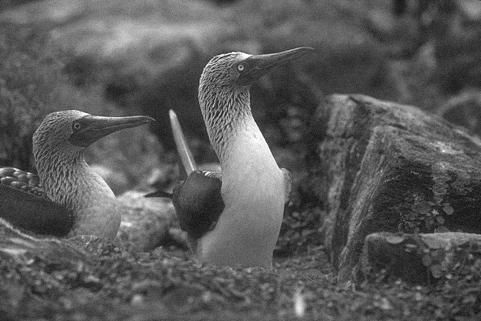}}
{\footnotesize (b) Corrupted $\mathbf{x}+\mathbf{n}_{o}$ (34.35dB/0.8985)}
\end{minipage}
\begin{minipage}[t]{0.24\textwidth}
\centering
\raisebox{-0.15cm}{\includegraphics[width=1\textwidth]{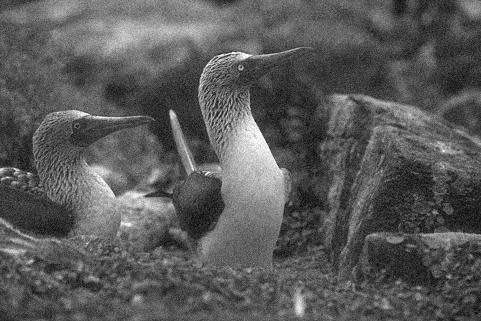}}
{\footnotesize (c) Doubly Corrupted $\mathbf{x}+\mathbf{n}_{o}+\mathbf{n}_{s}$ (28.00dB/0.6589)}
\end{minipage}
\begin{minipage}[t]{0.24\textwidth}
\centering
\raisebox{-0.15cm}{\includegraphics[width=1\textwidth]{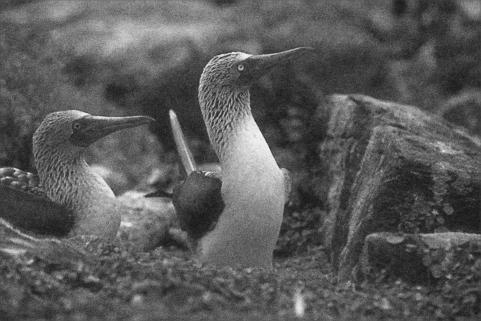}}
{\footnotesize (d) Output of DnCNN+NAC in training (31.51dB/0.8225)
}
\end{minipage}
}\vspace{-3mm}
\subfigure{
\begin{minipage}[t]{0.24\textwidth}
\centering
\raisebox{-0.15cm}{\includegraphics[width=1\textwidth]{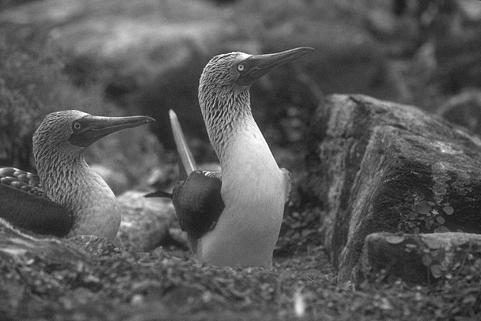}}
{\footnotesize (e) Output of DnCNN+NAC in test (\textbf{40.23}dB/\textbf{0.9663})}
\end{minipage}
\begin{minipage}[t]{0.24\textwidth}
\centering
\raisebox{-0.15cm}{\includegraphics[width=1\textwidth]{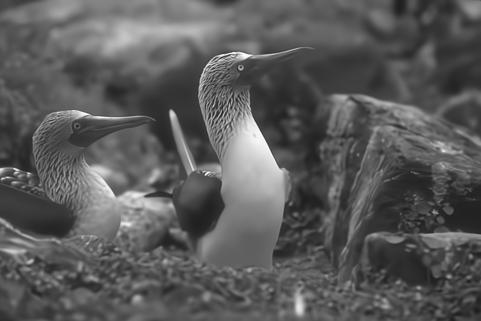}}
{\footnotesize (f) Denoised (b) by DnCNN~\cite{dncnn} (36.36dB/0.9384)}
\end{minipage}
\begin{minipage}[t]{0.24\textwidth}
\centering
\raisebox{-0.15cm}{\includegraphics[width=1\textwidth]{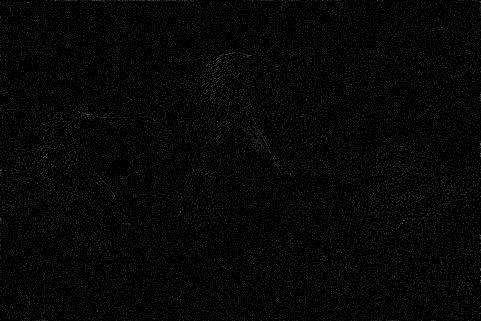}}
{\footnotesize (g) Estimated $\mathbf{n}_{s}$ \\ (difference between (c) and (d)) }
\end{minipage}
\begin{minipage}[t]{0.24\textwidth}
\centering
\raisebox{-0.15cm}{\includegraphics[width=1\textwidth]{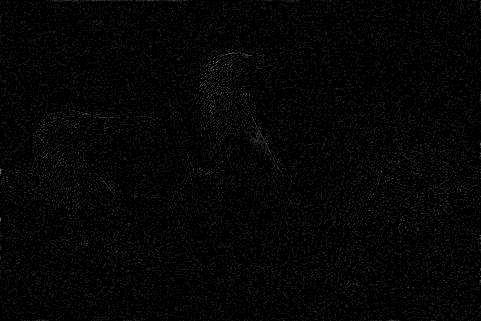}}
{\footnotesize (h) Estimated $\mathbf{n}_{o}$ \\ (difference between (b) and (e)) }
\end{minipage}
}\vspace{-1mm}
\caption{\textbf{An example to illustrate the pipeline of our NAC strategy based image denoising}.\
The image is ``\textsl{Test004}'' from the \textsl{BSD68} dataset.\
The observed noise $\mathbf{n}_{o}$ and simulated noise $\mathbf{n}_{s}$ are additive white Gaussian noise with $\sigma=5$.\
(a) The clean image $\mathbf{x}$.\ 
(b) The corrupted image $\mathbf{x}+\mathbf{n}_{o}$ (training target of our DnCNN+NAC).\ 
(c) The doubly corrutped image $\mathbf{x}+\mathbf{n}_{o}+\mathbf{n}_{s}$.\ 
(d) The output of training DnCNN in our NAC strategy, with input is the doubly corrupted image (c) and target is the corrupted image (b).\ 
(e) The output of our image-specific DnCNN+NAC tested on (b).\
PSNR and SSIM results of corresponding images are provided for objective references.
}
\label{fig:example}
\end{figure*}
Since the loss function $\mathcal{L}$ (usually $\ell_{2}$) and the conditional probability density functions $p(\mathbf{y}|\mathbf{x})$ and $p(\mathbf{z}|\mathbf{y})$ are all \textsl{continuous everywhere}, the optimal network parameters $\theta^{*}$ of Eqn.~(3) changes little with the addition of negligible noise $\mathbf{n}_{o}$ or $\mathbf{n}_{s}$.\ With Eqns.~(4)-(8), when the $\mathbf{x}$-conditioned expectation of $\mathbb{E}_{\mathbf{y}|\mathbf{x}}[\mathcal{L}(f_{\theta}(\mathbf{y}),\mathbf{x})]$ are replaced with the $\mathbf{y}$-conditioned expectation of $\mathbb{E}_{\mathbf{z}|\mathbf{y}}[\mathcal{L}(f_{\theta}(\mathbf{z}),\mathbf{y})]$, $f_{\theta}$ obtains similar $\mathbf{y}$-conditioned optimal parameters $\theta^{*}$:
\begin{equation}
\begin{split}
\label{e9}
&\arg\min_{\theta}\mathbb{E}_{\mathbf{y}}[\mathbb{E}_{\mathbf{z}|\mathbf{y}}[\mathcal{L}(f_{\theta}(\mathbf{z}),\mathbf{y})]]
\\
\approx
&\arg\min_{\theta}\mathbb{E}_{\mathbf{x}}[\mathbb{E}_{\mathbf{y}|\mathbf{x}}[\mathcal{L}(f_{\theta}(\mathbf{y}),\mathbf{x})]]
=
\theta^{*}.
\end{split}
\end{equation}
The network $f_{\theta}$ minimizes the loss function $\mathcal{L}$ for each input image pair separately, which equals to minimize it on all finite pairs of images.\ Through simple manipulations, Eqn.~(9) is equivalent to
\begin{equation}
\label{e10}
\begin{split}
&\arg\min_{\theta}\sum_{i=1}p(\mathbf{y}_{i})p(\mathbf{z}_{i}|\mathbf{y}_{i})\mathcal{L}(f_{\theta}(\mathbf{z}_{i}),\mathbf{y}_{i})
\\
=&
\arg\min_{\theta}\mathbb{E}_{\mathbf{y}}[\mathbb{E}_{\mathbf{z}|\mathbf{y}}[\mathcal{L}(f_{\theta}(\mathbf{z}),\mathbf{y})]]
\approx
\theta^{*}
.
\end{split}
\end{equation}
By exploring the dependence of
$p(\mathbf{z}_{i},\mathbf{y}_{i})=p(\mathbf{y}_{i})p(\mathbf{z}_{i}|\mathbf{y}_{i})$, Eqn.~(10) is equivalent to
\begin{equation}
\begin{split}
\label{e11}
&\arg\min_{\theta}\mathbb{E}_{(\mathbf{z},\mathbf{y})}[\mathcal{L}(f_{\theta}(\mathbf{z}),\mathbf{y})]
\\
=&
\arg\min_{\theta}\sum_{i=1}p(\mathbf{z}_{i},\mathbf{y}_{i})\mathcal{L}(f_{\theta}(\mathbf{z}_{i}),\mathbf{y}_{i})
\approx
\theta^{*}
.
\end{split}
\end{equation}

\noindent
\textbf{A simple but useful observation is}: \textsl{as long as the noise is weak, the optimal parameters of self-supervised network trained on noisy image pairs $\{(\mathbf{z}_{i},\mathbf{y}_{i})\}$ are very close to the optimal parameters of the supervised networks trained on pairs of noisy and clean images $\{(\mathbf{y}_{i},\mathbf{x}_{i})\}$}.\
In Figure~\ref{fig:example}, we explain our NAC strategy and illustrate this observation through an example on the image ``Test004'' from the BSD68 dataset:
The clean image $\mathbf{x}$ in (a) is firstly corrupted by observed AWGN noise $\mathbf{n}_{o}$ with $\sigma=5$.\ 
Then we add simulated AWGN noise $\mathbf{n}_{s}$ also with $\sigma=5$ to the corrupted image $\mathbf{x}+\mathbf{n}_{o}$ in (b), and obtain a doubly corrupted image $\mathbf{x}+\mathbf{n}_{o}+\mathbf{n}_{s}$ in (c).
The DnCNN~\cite{dncnn} with our NAC strategy, named as DnCNN+NAC, is trained with the doubly corrupted image $\mathbf{x}+\mathbf{n}_{o}+\mathbf{n}_{s}$ in (c) as input and the corrupted image $\mathbf{x}+\mathbf{n}_{o}$ in (b) as target.\ 
The final training output is plotted in (d), with similar PSNR and SSIM~\cite{ssim} results as the corrupted image $\mathbf{x}+\mathbf{n}_{o}$ in (b).\ 
Then the DnCNN+NAC network learned on (b) and (c) is directly employed to perform inference on the corrupted image $\mathbf{x}+\mathbf{n}_{o}$ in (b), and produces the testing output in (e).\ 
When compared to DnCNN~\cite{dncnn}, our DnCNN+NAC achieves much higher PSNR and SSIM results on the corrupted image (b). %
The estimated simulated noise $\mathbf{n}_{s}$ and observed noise $\mathbf{n}_{o}$ in training and test stages are plotted in (g) and (h), respectively.
One can see that they are visually in similar noise statistics. 


\noindent
\textbf{Consistency of noise statistics}.\ 
Since our contexts are the real-world scenarios, the noise can be modeled by mixed Poisson and Gaussian distribution~\cite{poissongaussian}.\ Fortunately, both the two distributions are linear additive, i.e., the addition variable of two Poisson (or Gaussian) distributed variables are still Poisson (or Gaussian) distributed.\ Assume that the observed (simulated) noise $\mathbf{n}_{o}$ ($\mathbf{n}_{s}$) follows a mixed $\mathbf{x}$-dependent ($\mathbf{y}$-dependent) Poisson distribution parameterized by $\lambda_{o}$ ($\lambda_{s}$) and Gaussian distribution $\mathcal{N}(\bm{0}, \sigma_{o}^{2})$ ($\mathcal{N}(\bm{0}, \sigma_{s}^{2})$), i.e.,
\begin{equation}
\begin{split}
\mathbf{n}_{o}
&\sim 
\mathbf{x}\odot \mathcal{P}(\lambda_{o})+\mathcal{N}(\bm{0}, \sigma_{o}^{2})
,
\\ 
\mathbf{n}_{s}
&\sim 
\mathbf{y}\odot \mathcal{P}(\lambda_{s})+\mathcal{N}(\bm{0}, \sigma_{s}^{2})
\\
&\approx
\mathbf{x}\odot \mathcal{P}(\lambda_{s})+\mathcal{N}(\bm{0}, \sigma_{s}^{2})
,
\end{split}
\end{equation}
where $\mathbf{x}\odot \mathcal{P}(\lambda_{o})$ and $\mathbf{y}\odot P(\lambda_{s})$ indicates that the noise $\mathbf{n}_{o}$ and $\mathbf{n}_{s}$ are element-wisely dependent on $\mathbf{x}$ and $\mathbf{y}$, respectively.\ The ``$\approx$'' is valid if we assume that the observed noise $\mathbf{n}_{o}$ is ``weak'' when compared to the signal $\mathbf{x}$.\
To this end, we have
\begin{equation}
\label{e13}
\mathbf{n}_{o}+\mathbf{n}_{s}
\sim 
\mathbf{x}\odot \mathcal{P}(\lambda_{o}+\lambda_{s})+\mathcal{N}(\bm{0}, \sigma_{o}^{2}+\sigma_{s}^{2}+2\rho\sigma_{o}\sigma_{s})
,
\end{equation}
where $\rho$ is the correlation between $\mathbf{n}_{o}$ and $\mathbf{n}_{s}$ ($\rho=0$ if they are independent).\ This indicates that the summed noise variable $\mathbf{n}_{o}+\mathbf{n}_{s}$ still follows a mixed $\mathbf{x}$ dependent Poisson and Gaussian distribution, guaranteeing the consistency in noise statistics between the \textsl{observed} realistic noise and the \textsl{simulated} noise.\ As will be validated by the experiments (\S\ref{sec:exp}), this property makes our NAC strategy consistently effective on different noise removal tasks.
\begin{table*}[t]
\small
\centering
\caption{\textbf{Average PSNR (dB) and SSIM~\cite{ssim} results of different methods on~\textsl{Set12} dataset} corrupted by AWGN noise.\ The first, second, and third best results are highlighted in \textcolor{red}{\textbf{red}}, \textcolor{blue}{\textbf{blue}}, and \textbf{bold}, respectively.}
\begin{tabular}{r||c|c|c|c|c|c|c|c|c|c}
\Xhline{1pt}
\rowcolor[rgb]{.85, .9, 0.95}
\multicolumn{1}{c||}{Noise Level}& \multicolumn{2}{c|}{$\sigma=5$ }& \multicolumn{2}{c|}{$\sigma=10$} & \multicolumn{2}{c|}{$\sigma=15$} & \multicolumn{2}{c|}{$\sigma=20$} & \multicolumn{2}{c}{$\sigma=25$} 
\\
\hline
\rowcolor[rgb]{.85, .9, 0.95}
\multicolumn{1}{c||}{Metric} & PSNR$\uparrow$ & SSIM$\uparrow$ & PSNR$\uparrow$ & SSIM$\uparrow$ & PSNR$\uparrow$ & SSIM$\uparrow$ & PSNR$\uparrow$ & SSIM$\uparrow$ & PSNR$\uparrow$ & SSIM$\uparrow$ 
\\
\hline
\hline
\textbf{BM3D}~\cite{bm3d} & 38.07 & 0.9580 & 34.40 & 0.9234 & 32.38 & 0.8957 & 31.00 & 0.8717 & 29.97 & 0.8503 \\ 
\textbf{DnCNN}~\cite{dncnn} & 38.76 & 0.9633 & 34.78 & 0.9270 & 32.86 & 0.9027 & 31.45 & 0.8799 & 30.43 & 0.8617 \\ 
\textbf{N2N}~\cite{noise2noise} & \textbf{39.72} & 0.9665 & 36.18 & 0.9446 & 33.99 & 0.9149 & \textbf{32.10} & 0.8788 & 30.72 & 0.8446 \\
\textbf{DIP}~\cite{dip} & 32.49 & 0.9344 & 31.49 & 0.9299 & 29.59 & 0.8636 & 27.67 & 0.8531 & 25.82 & 0.7723 \\
\textbf{N2V}~\cite{noise2void} & 27.06 & 0.8174 & 26.79 & 0.7859 & 26.12 & 0.7468 & 25.89 & 0.7405 & 25.01 & 0.6564 \\
\hline
\hline
\textbf{DnCNN+NAC} & \textcolor{red}{\textbf{43.17}} & \textcolor{blue}{\textbf{0.9817}} & \textcolor{red}{\textbf{37.16}} & \textbf{0.9336} & 33.64 & 0.8697 & 31.15 & 0.8024 & 29.22 & 0.7382 \\
\textbf{Blind DnCNN+NAC} & \textcolor{red}{\textbf{43.16}} & \textcolor{blue}{\textbf{0.9817}} & \textcolor{red}{\textbf{37.14}} & \textbf{0.9333} & 33.63 & 0.8693 & 31.14 & 0.8018 & 29.21 & 0.7376\\
\hline
\hline
\textbf{ResNet+NAC} &
\textcolor{blue}{\textbf{39.99}} & \textcolor{red}{\textbf{0.9820}} & \textbf{36.55} & \textcolor{red}{\textbf{0.9569}} & \textcolor{blue}{\textbf{34.24}} & \textcolor{red}{\textbf{0.9277}} & \textcolor{blue}{\textbf{32.46}} & \textcolor{blue}{\textbf{0.8961}} & \textcolor{blue}{\textbf{31.08}} & \textcolor{blue}{\textbf{0.8654}} \\
\textbf{Blind ResNet+NAC} & 38.48 & \textbf{0.9805} & \textcolor{blue}{\textbf{36.65}}	& \textcolor{blue}{\textbf{0.9564}} & \textcolor{red}{\textbf{34.77}}	& \textcolor{blue}{\textbf{0.9275}} & \textcolor{red}{\textbf{33.13}}	& \textcolor{red}{\textbf{0.9024}} & \textcolor{red}{\textbf{31.78}}	& \textcolor{red}{\textbf{0.8802}} \\
\hline
\end{tabular}
\vspace{-2mm}
\label{t-g12}
\end{table*}

\section{Learning Self-supervised Denoising Networks \\ 
by ``Noisy-As-Clean''}
\label{sec:self-sup}
Here, we propose to learn self-supervised denoising networks with our ``Noisy-As-Clean'' (NAC) strategy.\ %
We employ the DnCNN~\cite{dncnn} and ResNet in DIP~\cite{dip} as our baseline, and call the self-supervised networks as DnCNN+NAC and ResNet+NAC, respectively.\
Note that we only need the \textsl{observed} noisy image $\mathbf{y}$ to generate noisy image pairs $\{(\mathbf{z},\mathbf{y})\}$ with \textsl{simulated} noise $\mathbf{n}_{s}$, as illustrated in Figure~\ref{f1}.

\noindent
\textbf{Training self-supervised networks by our NAC}.
For real-world images captured by camera sensors, one can hardly distinguish the realistic noise from the signal.\ The signal intensity $\mathbf{x}$ is usually stronger than the noise intensity.\ That is, the expectation of the observed realistic noise $\mathbf{n}_{o}$ is usually much smaller than that of the latent clean image $\mathbf{x}$.\ If we train an image-specific network for the new noisy image $\mathbf{z}$ and regard the original noisy image $\mathbf{y}$ as the ground-truth image, then the trained image-specific network basically joint learn the image-specific prior and noise statistics.\ It has the capacity to remove the noise $\mathbf{n}_{s}$ from the new noisy image $\mathbf{z}$. Then if we perform denoising on the original noisy image $\mathbf{y}$, the observed noise $\mathbf{n}_{o}$ can be well-removed.\ Note that we \textsl{do not} use the clean image $\mathbf{x}$ as ``ground-truth'' in training the DnCNN+NAC and ResNet+NAC networks.\ 

\noindent
\textbf{Training blind denoising networks}.\ Most of existing supervised denoising networks train a specific model to process a fixed noise pattern~\cite{crosschannel2016,nlrn2018,upi}.\ To tackle unknown noise, one feasible solution for these networks is to assume the noise as AWGN and estimate its noise deviation.\ The corresponding noise is removed by using the networks trained with the estimated level.\ But this strategy largely degrades the denoising performance when the noise deviation is not estimated accurately.\ Besides, this solution can hardly deal with realistic noise, which is usually not AWGN, captured on real photographs.\ In order to be effective on removing realistic noise, the self-supervised networks by our NAC are feasible to blindly remove the unknown noise from real photographs.\ Inspired by~\cite{dncnn,cbdnet}, we propose to train a blind version of DnCNN+NAC and ResNet+NAC networks by using the AWGN noise within a range of levels (e.g., $[0, 55]$) for removing unknown AWGN noise.\ We also train blind ResNet+NAC with mixed AWGN and Poisson noise (both within a range of intensities) for removing the realistic noise.\ More details will be explained in \S\ref{sec:blind}.

\begin{table*}[htp]
\vspace{0mm}
\small
\centering
\caption{\textbf{Average PSNR (dB) and SSIM~\cite{ssim} results of different methods on \textsl{BSD68} dataset} corrupted by AWGN noise.\ The first, second, and third best results are highlighted in \textcolor{red}{\textbf{red}}, \textcolor{blue}{\textbf{blue}}, and \textbf{bold}, respectively.}
\begin{tabular}{r||c|c|c|c|c|c|c|c|c|c}
\Xhline{1pt}
\rowcolor[rgb]{ .85,  .9,  0.95}
\multicolumn{1}{c||}{Noise Level} & \multicolumn{2}{c|}{$\sigma=5$}& \multicolumn{2}{c|}{$\sigma=10$} & \multicolumn{2}{c|}{$\sigma=15$} & \multicolumn{2}{c|}{$\sigma=20$} & \multicolumn{2}{c}{$\sigma=25$}\\
\hline
\rowcolor[rgb]{ .85,  .9,  0.95}
\multicolumn{1}{c||}{Metric} & PSNR$\uparrow$ & SSIM$\uparrow$ & PSNR$\uparrow$ & SSIM$\uparrow$ & PSNR$\uparrow$ & SSIM$\uparrow$ & PSNR$\uparrow$ & SSIM$\uparrow$ & PSNR$\uparrow$ & SSIM$\uparrow$ \\
\hline
\hline
\textbf{BM3D}~\cite{bm3d} & 37.59 & 0.9640 & 33.32 & 0.9163 & 31.07 & 0.8720 & 29.62 & 0.8342 & 28.57 & 0.8017 \\
\textbf{DnCNN}~\cite{dncnn} & 38.07 & \textcolor{blue}{\textbf{0.9695}} & 33.88 & \textcolor{blue}{\textbf{0.9270}} & 31.73 & 0.8706 & \textbf{30.27} & \textbf{0.8563} & \textcolor{blue}{\textbf{29.23}} & \textcolor{blue}{\textbf{0.8278}} \\
\textbf{N2N}~\cite{noise2noise} & \textbf{38.58} & 0.9627 & 34.07 & 0.9200 & \textbf{31.81} & \textbf{0.8770} & 30.14 & 0.8550 & 28.67 & 0.8123 \\ 
\textbf{DIP}~\cite{dip} & 29.74 & 0.8435 & 28.16 & 0.8310 & 27.07 & 0.7867 & 25.80 & 0.7205 & 24.63 & 0.6680 \\ 
\textbf{N2V}~\cite{noise2void} & 26.70 & 0.7915 & 26.39 & 0.7621 & 25.77 & 0.7126 & 25.41 & 0.6678 & 24.83 & 0.6305 \\
\hline
\hline
\textbf{DnCNN+NAC} & \textcolor{red}{\textbf{40.21}} & \textbf{0.9674} & \textbf{34.21} & 0.8913 & 30.72 & 0.8044 & 28.25 & 0.7230 & 26.34 & 0.6515 \\
\textbf{Blind DnCNN+NAC} & \textcolor{red}{\textbf{40.20}} & \textbf{0.9674} & \textbf{34.21} & 0.8911 & 30.71 & 0.8041 & 28.24 & 0.7227 & 26.33 & 0.6511\\
\hline
\hline
\textbf{ResNet+NAC} & \textcolor{blue}{\textbf{39.00}} & \textcolor{red}{\textbf{0.9707}} & \textcolor{red}{\textbf{34.60}} & \textcolor{red}{\textbf{0.9324}} & \textcolor{red}{\textbf{32.13}} & \textcolor{red}{\textbf{0.8942}} & \textcolor{blue}{\textbf{30.47}} & \textcolor{red}{\textbf{0.8636}} & \textbf{28.96} & \textbf{0.8185}\\
\textbf{Blind ResNet+NAC} & 38.26	&0.9605	& \textcolor{blue}{\textbf{34.26}} & \textbf{0.9266} & \textcolor{blue}{\textbf{32.06}} & \textcolor{blue}{\textbf{0.8919}} & \textcolor{red}{\textbf{30.50}}	& \textcolor{blue}{\textbf{0.8609}} & \textcolor{red}{\textbf{29.33}}	& \textcolor{red}{\textbf{0.8327}} \\
\hline
\end{tabular}
\vspace{-2.5mm}
\label{t-g68}
\end{table*}

\noindent
\textbf{Testing} is performed by directly regarding an \textsl{observed} noisy image $\mathbf{y}=\mathbf{x}+\mathbf{n}_{o}$ as input.\ 
We only test the image $\mathbf{y}$ once.\ 
The denoised image can be represented as $\hat{\mathbf{y}}=f_{\theta^{*}}(\mathbf{y})$, with which the objective metrics, e.g., PSNR and SSIM~\cite{ssim}, can be computed with the clean image $\mathbf{x}$.\

\noindent
\textbf{Implementation details}.\ We employ the DnCNN~\cite{dncnn} and ResNet in DIP~\cite{dip} as the backbones, and turn them into self-supervised networks by our NAC strategy, which are named as DnCNN+NAC and ResNet+NAC, respectively.
The DnCNN contains 17 layers of convolution, Batch Normalization (BN)~\cite{bn2015}, and Rectified Linear Units (ReLU) activation operator~\cite{relu2010}.\  
To accommodate DnCNN with our NAC strategy, we set the output of  DnCNN+NAC as the denoised image, not the residual noise in DnCNN~\cite{dncnn}.\ 
We observe no difference between the results on PSNR, SSIM~\cite{ssim}, and visual quality by employing these two types of outputs in our experiments.
As DnCNN, the parameters of DnCNN+NAC are initialized from a pretrained ResNet.
As used in~\cite{dip}, the ResNet in our ResNet+NAC includes $10$ residual blocks, each containing two convolutional layers followed by a BN~\cite{bn2015} and a ReLU~\cite{relu2010} after the first BN.\ 
The parameters are randomly initialized without being pretrained.\ 
For both baselines, the optimizer is Adam~\cite{adam} with default parameters.\ 
The learning rate is fixed at $0.001$ in all experiments.\ 
We use the $\ell_{2}$ loss function.\ 
For each test image, we only train the DnCNN+NAC in 100 epochs, while the original DnCNN is trained with 180 epochs.\ 
The ResNet+NAC is trained in $1000$ epochs for each test image, the same as that in DIP~\cite{dip}. 
As suggested by DnCNN~\cite{dncnn} and DIP~\cite{dip}, we employ $4$ rotations \{0\degree, 90\degree, 180\degree, 270\degree\} combined with 2 mirror (vertical and horizontal) reflections, resulting in totally $8$ transformations for data augmentation.\ 
We implement the DnCNN+NAC and ResNet+NAC networks in PyTorch.

\section{Experiments}
\label{sec:exp}

In this section, we evaluate the performance of our ``Noisy-As-Clean'' (NAC) networks on image denoising.\ In all experiments, we train a denoising network using only the noisy test image $\mathbf{y}$ as the target, and using the \textsl{simulated} noisy image $\mathbf{z}$ as the input.\ 
For all comparison methods, the source codes or trained models are downloaded from the corresponding authors' websites.\ We use the default parameter settings, unless otherwise specified.\
The PSNR, SSIM~\cite{ssim}, and visual quality of different methods are used to evaluate the comparison.\ 
We first test with synthetic noise such as additive white Gaussian noise (AWGN) in \S\ref{sec:syn}, continue to perform blind image denoising in \S\ref{sec:blind}, and finally tackle the realistic noise in \S\ref{sec:real}.\ 
In \S\ref{sec:dis}, we conduct comprehensive ablation studies to gain deeper insights into our NAC strategy.
\begin{table*}[t]
\centering
\caption{\textbf{Average PSNR (dB) and SSIM~\cite{ssim} of different methods} on the \textsl{CC} dataset~\cite{crosschannel2016} and the \textsl{DND} dataset~\cite{dnd2017}.\ The best results are highlighted in \textbf{bold}.\ ``NA'' means ``Not Available'' due to unavailable code (GCBD on \textsl{CC}~\cite{crosschannel2016}) or difficult experiments (DIP on \textsl{DND}~\cite{dnd2017}).\ 
The first, second, and third best results are highlighted in \textcolor{red}{\textbf{red}}, \textcolor{blue}{\textbf{blue}}, and \textbf{bold}, respectively.
}
\renewcommand\arraystretch{1}
\begin{tabular}{r||c|cc|cc|cc|cc}
\Xhline{1pt}
\rowcolor[rgb]{ .85,  .9,  0.95}
&
\multicolumn{1}{c|}{Type}&
\multicolumn{2}{c|}{Traditional Methods}&
\multicolumn{2}{c|}{Supervised Networks}& \multicolumn{2}{c|}{Unsupervised Networks}& \multicolumn{2}{c}{Self-supervised Networks}
\\
\cline{2-10}
\rowcolor[rgb]{ .85,  .9,  0.95}
\multicolumn{1}{c||}{\multirow{-2}{*}{Dataset}}
&
Method 
& \textbf{CBM3D}~\cite{cbm3d}&\textbf{NI}~\cite{neatimage}&\textbf{DnCNN+}\cite{dncnn}&\textbf{CBDNet}~\cite{cbdnet}&\textbf{GCBD}~\cite{gcbd}&\textbf{N2N}~\cite{noise2noise}&\textbf{DIP}~\cite{dip}&\textbf{Blind ResNet+NAC}
\\
\hline
\multicolumn{1}{r||}{\multirow{2}{*}{\textsl{CC}~\cite{crosschannel2016}}}
&PSNR$\uparrow$ & 35.19 & 35.33 & 35.40 & \textcolor{blue}{\textbf{36.44}} & NA & 35.32 & \textbf{35.69} & \textcolor{red}{\textbf{36.59}}
\\
&SSIM$\uparrow$ & 0.9063 & 0.9212 & 0.9115 & \textcolor{blue}{\textbf{0.9460}} & NA & 0.9160 & \textbf{0.9259} & \textcolor{red}{\textbf{0.9502}}
\\
\hline
\multicolumn{1}{c||}{\multirow{2}{*}{\textsl{DND}~\cite{dnd2017}}}
&PSNR$\uparrow$ & 34.51 & 35.11 & \textcolor{blue}{\textbf{37.90}} & \textcolor{red}{\textbf{38.06}} & 35.58 & 33.10 & NA & \textbf{36.20} 
\\
&SSIM$\uparrow$ & 0.8507 & 0.8778 & \textcolor{red}{\textbf{0.9430}} & \textcolor{blue}{\textbf{0.9421}} & 0.9217 & 0.8110 & NA & \textbf{0.9252} 
\\
\hline
\end{tabular}
\vspace{-2.5mm}
\label{t-cc+dnd}
\end{table*}
\begin{figure*}[htbp]
\centering
\subfigure{
\begin{minipage}[t]{0.24\textwidth}
\centering
\raisebox{-0.15cm}{\includegraphics[width=1\textwidth]{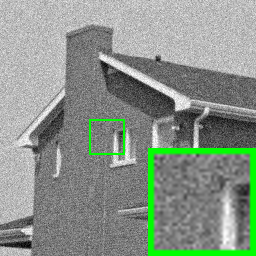}}
{\footnotesize (a) Noisy (24.59dB/0.4456)}
\end{minipage}
\begin{minipage}[t]{0.24\textwidth}
\centering
\raisebox{-0.15cm}{\includegraphics[width=1\textwidth]{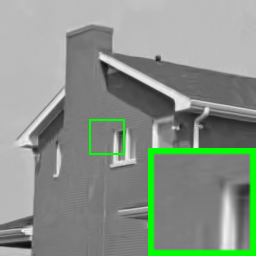}}
{\footnotesize (b) BM3D~\cite{bm3d} (34.93dB/0.8907)}
\end{minipage}
\begin{minipage}[t]{0.24\textwidth}
\centering
\raisebox{-0.15cm}{\includegraphics[width=1\textwidth]{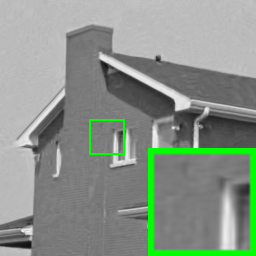}}
{\footnotesize (c) PGPD~\cite{pgpd} (34.83dB/0.8850)}
\end{minipage}
\begin{minipage}[t]{0.24\textwidth}
\centering
\raisebox{-0.15cm}{\includegraphics[width=1\textwidth]{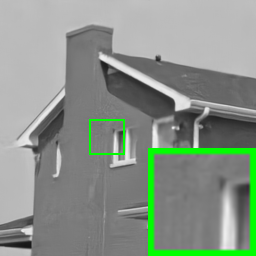}}
{\footnotesize (d) DnCNN~\cite{dncnn} (34.98dB/0.8846)}
\end{minipage}
}\vspace{-3mm}
\subfigure{
\begin{minipage}[t]{0.24\textwidth}
\centering
\raisebox{-0.15cm}{\includegraphics[width=1\textwidth]{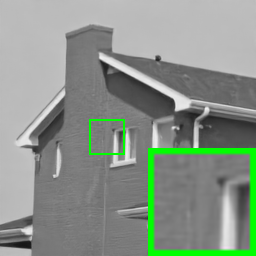}}
{\footnotesize (e) N2N~\cite{noise2noise} (35.74dB/0.9019)}
\end{minipage}
\begin{minipage}[t]{0.24\textwidth}
\centering
\raisebox{-0.15cm}{\includegraphics[width=1\textwidth]{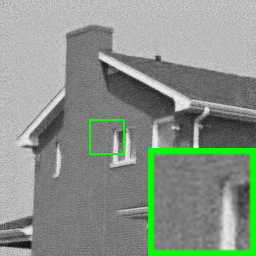}}
{\footnotesize (f) DIP~\cite{dip} (30.38dB/0.7145)}
\end{minipage}
\begin{minipage}[t]{0.24\textwidth}
\centering
\raisebox{-0.15cm}{\includegraphics[width=1\textwidth]{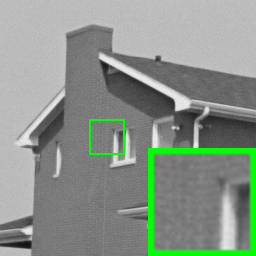}}
{\footnotesize (g) ResNet+NAC (\textbf{35.89}dB/\textbf{0.9101}) }
\end{minipage}
\begin{minipage}[t]{0.24\textwidth}
\centering
\raisebox{-0.15cm}{\includegraphics[width=1\textwidth]{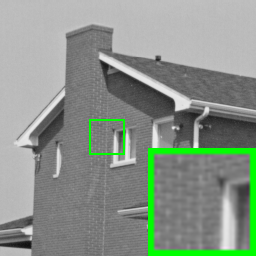}}
{\footnotesize (h) Ground Truth}
\end{minipage}
}\vspace{-1mm}
\caption{\textbf{Denoised images and PSNR/SSIM results of ``\textsl{House}'' in \textsl{Set12} by different methods}.\ The images are corrupted by AWGN noise with $\sigma=15$.\ The best results on PSNR and SSIM are highlighted in \textbf{bold}.}
\label{fig:awgn15}
\end{figure*}

\subsection{Synthetic Noise Removal With Known Noise}
\label{sec:syn}
We evaluate the DnCNN+NAC and ResNet+NAC networks on images corrupted by synthetic AWGN noise.\ 
More results on signal dependent Poisson noise and mixed Poisson-AWGN noise are provided in the \textsl{Supplementary File}.\

\noindent
\textbf{Training self-supervised networks}.\ Here, we train an image-specific denoising network using the \textsl{observed} noisy test image $\mathbf{y}$ as the target, and the \textsl{simulated} noisy image $\mathbf{z}$ as the input.\ Each \textsl{observed} noisy image $\mathbf{y}=\mathbf{x}+\mathbf{n}_{o}$ is generated by adding the \textsl{observed} noise $\mathbf{n}_{o}$ to the clean image $\mathbf{x}$.\ The \textsl{simulated} noisy image $\mathbf{z}=\mathbf{y}+\mathbf{n}_{s}$ is generated by adding \textsl{simulated} noise $\mathbf{n}_{s}$ to \textsl{observed} noisy image $\mathbf{y}$.\

\noindent
\textbf{Comparison methods}.\
We compare DnCNN+NAC and ResNet+NAC networks with state-of-the-art image denoising methods~\cite{bm3d,dncnn,noise2noise}.\ On AWGN noise, we compare with BM3D~\cite{bm3d}, DnCNN~\cite{dncnn}, Noise2Noise (N2N)~\cite{noise2noise}, Deep Image Prior (DIP)~\cite{dip}, and Noise2Void (N2V)~\cite{noise2void}.\ 

\noindent
\textbf{Test datasets}.\ We evaluate the comparison methods on the \textsl{Set12} and \textsl{BSD68} datasets, which are widely tested by supervised denoising networks~\cite{dncnn,nlrn2018} and previous methods~\cite{bm3d,pgpd}.\ The \textsl{Set12} dataset contains 12 images of sizes $512\times512$ or $256\times256$, while the \textsl{BSD68} dataset contains 68 images of different sizes.

\begin{figure*}[htbp]
\centering
\subfigure{
\begin{minipage}[t]{0.24\textwidth}
\centering
\raisebox{-0.15cm}{\includegraphics[width=1\textwidth]{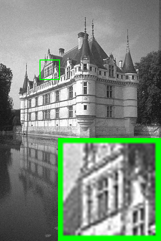}}
{\footnotesize (a) Noisy (34.15dB/0.8416)}
\end{minipage}
\begin{minipage}[t]{0.24\textwidth}
\centering
\raisebox{-0.15cm}{\includegraphics[width=1\textwidth]{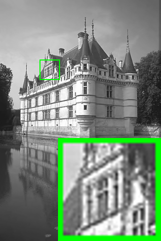}}
{\footnotesize (b) BM3D~\cite{bm3d} (38.20dB/0.9569)}
\end{minipage}
\begin{minipage}[t]{0.24\textwidth}
\centering
\raisebox{-0.15cm}{\includegraphics[width=1\textwidth]{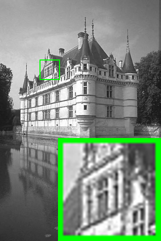}}
{\footnotesize (c) PGPD~\cite{pgpd} (38.02dB/0.9524)}
\end{minipage}
\begin{minipage}[t]{0.24\textwidth}
\centering
\raisebox{-0.15cm}{\includegraphics[width=1\textwidth]{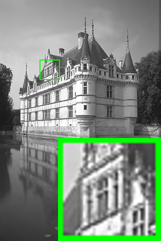}}
{\footnotesize (d) DnCNN~\cite{dncnn} (38.64dB/0.9559)}
\end{minipage}
}\vspace{-3mm}
\subfigure{
\begin{minipage}[t]{0.24\textwidth}
\centering
\raisebox{-0.15cm}{\includegraphics[width=1\textwidth]{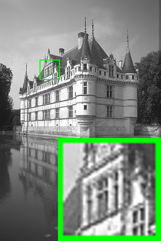}}
{\footnotesize (e) N2N~\cite{noise2noise} (39.63dB/0.9682)}
\end{minipage}
\begin{minipage}[t]{0.24\textwidth}
\centering
\raisebox{-0.15cm}{\includegraphics[width=1\textwidth]{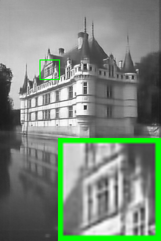}}
{\footnotesize (f) DIP~\cite{dip} (27.22dB/0.8794)}
\end{minipage}
\begin{minipage}[t]{0.24\textwidth}
\centering
\raisebox{-0.15cm}{\includegraphics[width=1\textwidth]{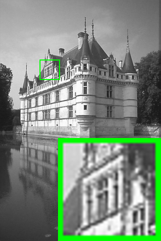}}
{\footnotesize (g) ResNet+NAC (\textbf{39.89}dB/\textbf{0.9693}) }
\end{minipage}
\begin{minipage}[t]{0.24\textwidth}
\centering
\raisebox{-0.15cm}{\includegraphics[width=1\textwidth]{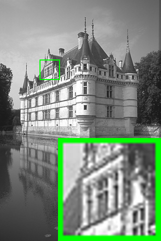}}
{\footnotesize (h) Ground Truth}
\end{minipage}
}\vspace{-1mm}
\caption{\textbf{Denoised images and PSNR/SSIM results of ``\textsl{Test003}'' in \textsl{BSD68} by different methods}.\ The images are corrupted by AWGN noise with $\sigma=5$.\ The best results on PSNR and SSIM are highlighted in \textbf{bold}.
}
\label{fig:awgn5}
\end{figure*}

\noindent
\textbf{Results on AWGN noise} with noise levels (standard deviation, or std) of $\sigma\in\{5, 10, 15, 20, 25\}$ are provided here.\ 
The \textsl{observed} noise $\mathbf{n}_{o}$ is AWGN with std of $\sigma$, while the \textsl{simulated} noise $\mathbf{n}_{s}$ is with the same $\sigma$ as that of $\mathbf{n}_{o}$.\ 
The comparison results are listed in Tables~\ref{t-g12} and~\ref{t-g68}.\ 
It can be seen that, DnCNN+NAC achieves better PSNR and SSIM results than those of the original DnCNN when $\sigma=5,10$. 
Note that DnCNN are supervised networks trained offline on the \textsl{BSD400} dataset, while the variant DnCNN+NAC network is trained online for each corrupted image.\
Besides, the blind version of DnCNN+NAC achieves negligible performance drop when compared to the DnCNN+NAC, which is consistent with~\cite{dncnn}.\ 
On the other side, the ResNet+NAC networks achieve comparable or better performance on PSNR and SSIM~\cite{ssim} than BM3D~\cite{bm3d} and DnCNN~\cite{dncnn}, especially when the noise levels are weak ($\sigma=5,10$).\ 
Besides, our ResNet+NAC networks outperform the other unsupervised and self-supervised networks such as N2N~\cite{noise2noise}, DIP~\cite{dip}, and N2V~\cite{noise2void} by a large margin on PSNR and SSIM~\cite{ssim}.\
In Figures~\ref{fig:awgn15} and~\ref{fig:awgn5}, we provide the visual comparisons of the denoised images by the competing methods.\ 
One can see that the ResNet+NAC networks produce better image quality and higher PSNR/SSIM results than the comparison methods.


\subsection{Synthetic Noise Removal With Unknown Noise}
\label{sec:blind}

To deal with unknown noise, we propose to train blind versions of the DnCNN~\cite{dncnn} and ResNet in~\cite{dip} by our NAC strategy.\ 
Here, we test the Blind DnCNN+NAC and Blind ResNet+NAC networks on AWGN noise with unknown noise deviation.\ We use the same training strategy, comparison methods, and test datasets as in \S\ref{sec:syn}.

\noindent
\textbf{Training blind networks}.\ We train the Blind DnCNN+NAC and Blind ResNet+NAC networks on the corrupted test image degraded \textsl{again} by AWGN noise with unknown noise levels (deviations).\
The noise levels are randomly sampled in Gaussian distribution within $[0, 55]$.\ We also test on noise levels in uniform distribution and obtain similar results.\ We repeat the training of DnCNN+NAC and ResNet+NAC networks on the test image with different deviations.\ 

\noindent
\textbf{Results on blind denoising}.\ For the same test image, we add to it the AWGN noise whose deviation is also in $\{5, 10, 15, 20, 25\}$.\ The blindly trained DnCNN+NAC and ResNet+NAC networks are directly utilized to denoise the test image without estimating its deviation.\ The results are also listed in Tables~\ref{t-g12} and~\ref{t-g68}.\ We observe that, the Blind ResNet+NAC networks trained on AWGN noise with unknown levels can achieve even better PSNR and SSIM~\cite{ssim} results than the ResNet+NAC networks trained on specific noise levels.\ Note that on \textsl{BSD68}, the ResNet+NAC networks achieve higher PSNR and SSIM results than DnCNN~\cite{dncnn}.\ This demonstrates the effectiveness of our ResNet+NAC networks on blind image denoising.\ With the success on blind image, next we will turn to real-world image denoising, in which the noise is also unknown and very complex. 

\subsection{Practice on Real Photographs}
\label{sec:real}
With the promising performance on blind image denoising, here we tackle the realistic noise for practical applications.\ The \textsl{observed} realistic noise $\mathbf{n}_{o}$ can be roughly modeled as mixed Poisson noise and AWGN noise~\cite{poissongaussian,cbdnet}.\ Hence, for each \textsl{observed} noisy image $\mathbf{y}$, we generate the \textsl{simulated} noise $\mathbf{n}_{s}$ by sampling the $\mathbf{y}$-dependent Poisson part and the independent AWGN noise.\
\begin{figure*}
\centering
\subfigure{
\begin{minipage}[t]{0.32\textwidth}
\centering
\raisebox{-0.15cm}{\includegraphics[width=1\textwidth]{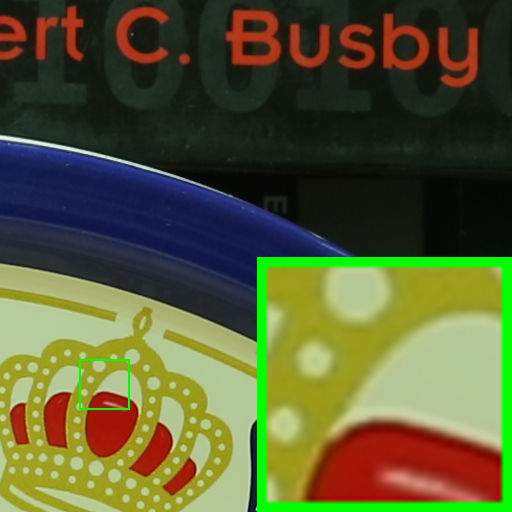}}
{\footnotesize (a) Ground Truth}
\end{minipage}
\begin{minipage}[t]{0.32\textwidth}
\centering
\raisebox{-0.15cm}{\includegraphics[width=1\textwidth]{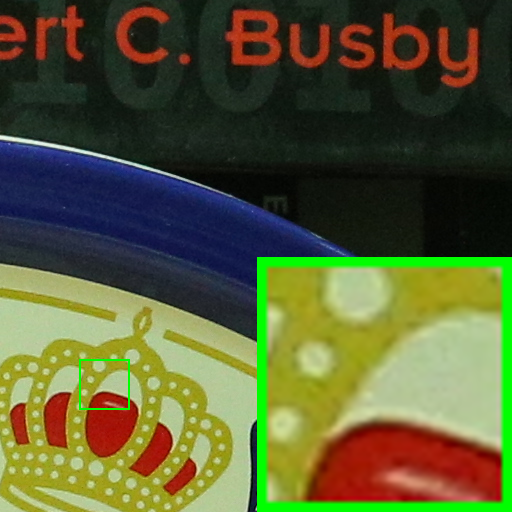}}
{\footnotesize (b) Noisy (36.25dB/0.9345)}
\end{minipage}
\begin{minipage}[t]{0.32\textwidth}
\centering
\raisebox{-0.15cm}{\includegraphics[width=1\textwidth]{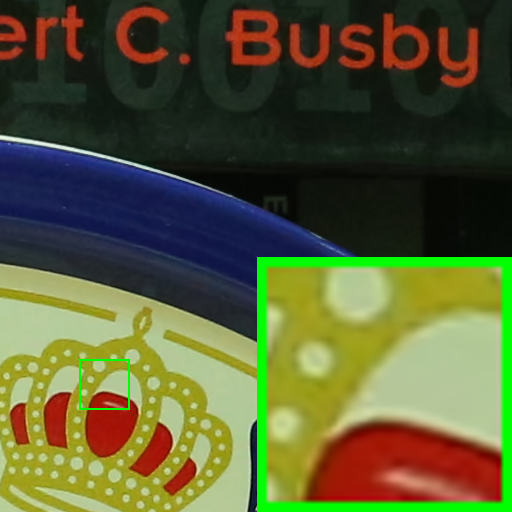}}
{\footnotesize (c) CBM3D~\cite{cbm3d} (36.61dB/0.9669)}
\end{minipage}
}\vspace{-3mm}
\subfigure{
\begin{minipage}[t]{0.32\textwidth}
\centering
\raisebox{-0.15cm}{\includegraphics[width=1\textwidth]{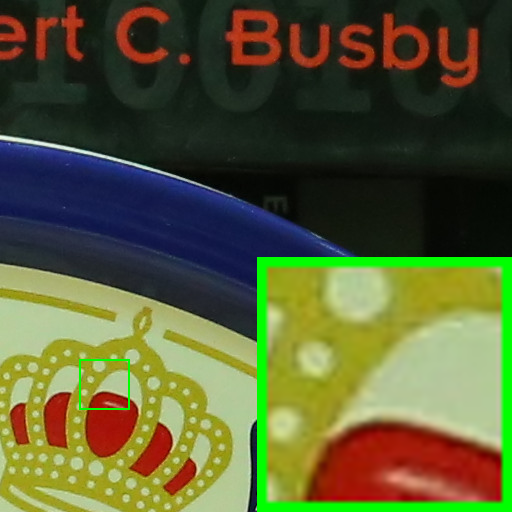}}
{\footnotesize (d) NI~\cite{neatimage} (37.58dB/0.9600)}
\end{minipage}
\begin{minipage}[t]{0.32\textwidth}
\centering
\raisebox{-0.15cm}{\includegraphics[width=1\textwidth]{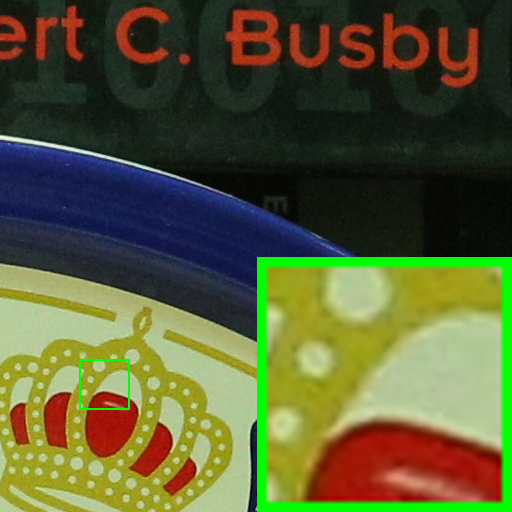}}
{\footnotesize (e) DnCNN+~\cite{dncnn} (37.16dB/0.9389)}
\end{minipage}
\begin{minipage}[t]{0.32\textwidth}
\centering
\raisebox{-0.15cm}{\includegraphics[width=1\textwidth]{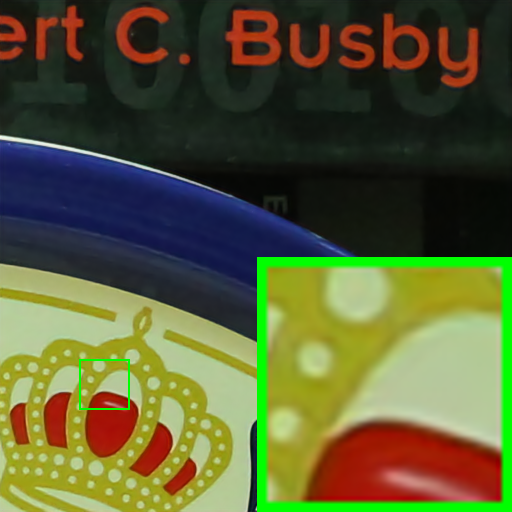}}
{\footnotesize (f) CBDNet~\cite{cbdnet} (36.58dB/0.9613) }
\end{minipage}
}\vspace{-3mm}
\subfigure{
\begin{minipage}[t]{0.32\textwidth}
\centering
\raisebox{-0.15cm}{\includegraphics[width=1\textwidth]{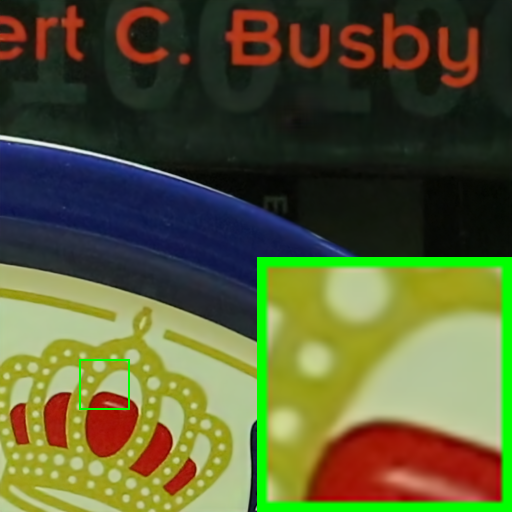}}
{\footnotesize (g) N2N~\cite{noise2noise} (36.99dB/0.9604)}
\end{minipage}
\begin{minipage}[t]{0.32\textwidth}
\centering
\raisebox{-0.15cm}{\includegraphics[width=1\textwidth]{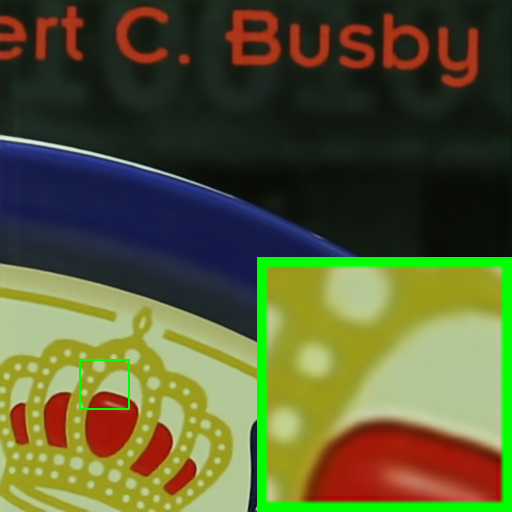}}
{\footnotesize (h) DIP~\cite{dip} (35.99dB/0.9529)}
\end{minipage}
\begin{minipage}[t]{0.32\textwidth}
\centering
\raisebox{-0.15cm}{\includegraphics[width=1\textwidth]{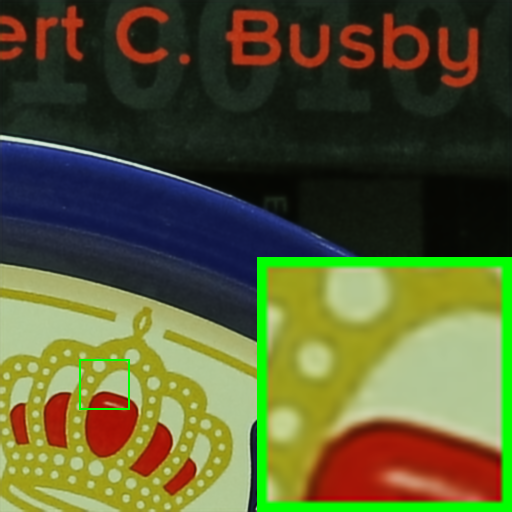}}
{\footnotesize (i) Blind ResNet+NAC (\textbf{37.88}dB/\textbf{0.9729})}
\end{minipage}
}\vspace{-1mm}
\caption{\textbf{Denoised images and PSNR/SSIM results of ``\textsl{5dmark3-iso3200-1}'' in the \textsl{Cross-Channel} dataset~\cite{crosschannel2016} by different methods}.\ The best results are highlighted in \textbf{bold}.}
\label{fig:cc}
\end{figure*}

\noindent
\textbf{Training blind ResNet+NAC networks} is also performed for each test image, i.e., the \textsl{observed} noisy image $\mathbf{y}$.\ In real-world scenarios, each \textsl{observed} noisy image $\mathbf{y}$ is corrupted without knowing the specific noise statistics of the \textsl{observed} noise $\mathbf{n}_{o}$.\ Therefore, the \textsl{simulated} noise $\mathbf{n}_{s}$ is directly estimated on $\mathbf{y}$ as mixed $\mathbf{y}$-dependent Poisson and AWGN noise.\ For each transformation image in data augmentation, the Poisson noise is randomly sampled with the parameter $\lambda$ in $0<\lambda\le25$, and the AWGN noise is randomly sampled with the noise level $\sigma$ in $0<\sigma\le25$.

\noindent
\textbf{Comparison methods}.\
We compare with state-of-the-art methods on real-world image denoising, including CBM3D~\cite{cbm3d}, the commercial software Neat Image~\cite{neatimage}, two supervised networks DnCNN+~\cite{dncnn} and CBDNet~\cite{cbdnet}, and two unsupervised networks GCBD~\cite{gcbd} and Noise2Noise~\cite{noise2noise}, and the self-supervised network DIP~\cite{dip}.\ Note that DnCNN+~\cite{dncnn} and CBDNet~\cite{cbdnet} are two state-of-the-art supervised networks for real-world image denoising, and DnCNN+ is an improved extension of DnCNN~\cite{dncnn} with better performance (the authors of DnCNN+ provide us the models/results of DnCNN+).

\begin{figure*}[ht!]
\centering
\begin{minipage}[t]{0.245\textwidth}
\includegraphics[width=1\textwidth]{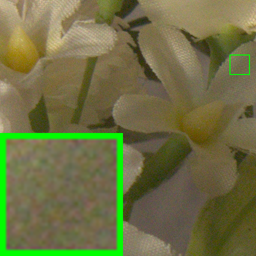}
\centering{\scriptsize (a) \scriptsize Noisy: 31.46dB/0.9370}
\includegraphics[width=1\textwidth]{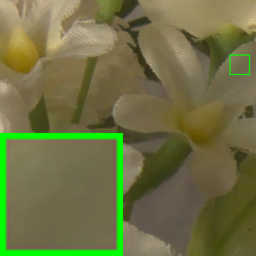}
\centering{\scriptsize (e) \scriptsize CBDNet~\cite{cbdnet}: 39.34dB/0.9905}
\end{minipage}
\begin{minipage}[t]{0.245\textwidth}
\includegraphics[width=1\textwidth]{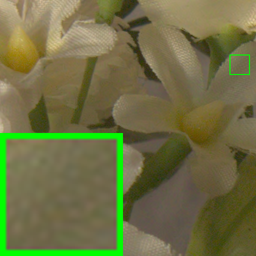}
\centering{\scriptsize (b) \scriptsize CBM3D~\cite{cbm3d}: 36.26dB/0.9811}
\includegraphics[width=1\textwidth]{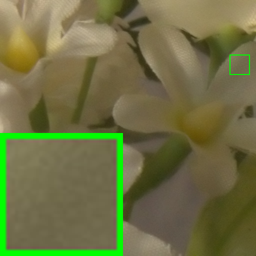}
\centering{\scriptsize (f) \scriptsize GCBD~\cite{gcbd}: 37.52dB/0.9765}
\end{minipage}
\begin{minipage}[t]{0.245\textwidth}
\includegraphics[width=1\textwidth]{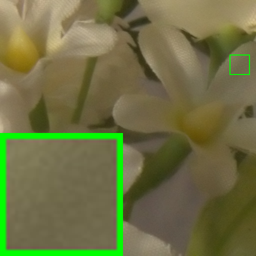}
\centering{\scriptsize (c) \scriptsize NI~\cite{neatimage}: 37.52dB/0.9868}
\includegraphics[width=1\textwidth]{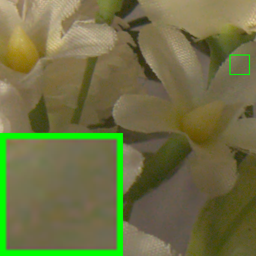}
\centering{\scriptsize (g) \scriptsize N2N~\cite{noise2noise}: 34.95dB/0.9621}
\end{minipage}
\begin{minipage}[t]{0.245\textwidth}
\includegraphics[width=1\textwidth]{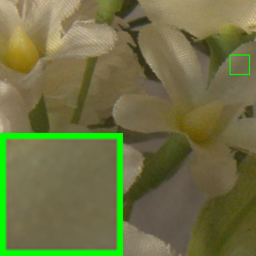}
\centering{\scriptsize (d) \scriptsize DnCNN+~\cite{dncnn}: 38.25dB/0.9888}
\includegraphics[width=1\textwidth]{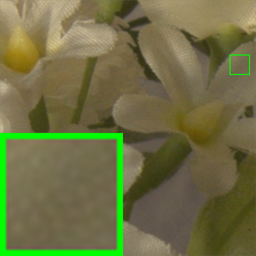}
\centering{\scriptsize (h) \scriptsize Blind ResNet+NAC: 38.34dB/0.9887}
\end{minipage}
\vspace{0mm}
\caption{\textbf{Denoised images and PSNR(dB)/SSIM by comparison methods} on ``\textsl{0017\_3}'' in \textsl{DND}~\cite{dnd2017}.\ The ``ground-truth'' image is not released, but PSNR(dB)/SSIM results are publicly provided on \href{https://noise.visinf.tu-darmstadt.de/benchmark/\#results_srgb}{\textsl{DND} Benchmark}.}
\vspace{-2.5mm}
\label{fig:dnd}
\end{figure*}

\noindent
\textbf{Test datasets}.\ We evaluate the comparison methods on the \textsl{Cross-Channel} (\textsl{CC}) dataset~\cite{crosschannel2016} and \textsl{DND} dataset~\cite{dnd2017}.\
The \textsl{CC} dataset~\cite{crosschannel2016} includes noisy images of 11 static scenes captured by Canon 5D Mark 3, Nikon D600, and Nikon D800 cameras.\ The noisy images are collected under a highly controlled indoor environment.\ Each scene is shot $500$ times using the same settings.\ The average of the $500$ shots is taken as ``ground-truth''.\ We use the default $15$ images of size $512\times512$ cropped by the authors to evaluate different image denoising methods.\ 
The \textsl{DND} dataset \cite{dnd2017} contains 50 scenarios captured by Sony A7R, Olympus E-M10, Sony RX100 IV, and Huawei Nexus 6P.\ Each scene is cropped to 20 bounding boxes of $512\times512$ pixels, generating totally 1000 test images.\ The noisy images are collected under higher ISO values with shorter exposure times, while the ``ground truth'' images are captured under lower ISO values with adjusted longer exposure times.\ The ``ground truth'' images are not released, but we can obtain the PSNR and SSIM results by submitting the denoised images to the \href{https://noise.visinf.tu-darmstadt.de/benchmark/\#results_srgb}{\textsl{DND}'s Website}.\

\noindent
\textbf{Comparison results on PSNR and SSIM} are listed in Table~\ref{t-cc+dnd}.\ 
As can be seen, the ResNet+NAC networks achieve better performance than all previous denoising methods, including the CBM3D~\cite{cbm3d}, the supervised networks DnCNN+~\cite{dncnn} and CBDNet~\cite{cbdnet}, and the unsupervised networks GCBD~\cite{gcbd}, N2N~\cite{noise2noise}, and DIP~\cite{dip}.\ This demonstrates that the ResNet+NAC networks can indeed handle the complex, unknown, and realistic noise, and achieve better performance than supervised networks such as DnCNN+~\cite{dncnn} and CBDNet~\cite{cbdnet}.\

\noindent
\textbf{Qualitative results}.\ 
In Figures~\ref{fig:cc} and~\ref{fig:dnd}, we show the denoised images of our ResNet+NAC and the comparison methods on the images of ``\textsl{5dmark3-iso3200-1}'' from the \textsl{CC} dataset~\cite{crosschannel2016} and ``\textsl{0017\_3}'' from the \textsl{DND} dataset~\cite{dnd2017}, respectively.\ We observe that our self-supervised Blind ResNet+NAC is very effective on removing realistic noise from the real photograph.\ Besides, the Blind ResNet+NAC networks achieve competitive PSNR and SSIM results when compared with the other methods, including the supervised DnCNN+~\cite{dncnn} and CBDNet~\cite{cbdnet}. 

\noindent
\textbf{Speed}.\ The work most similar to ours is Deep Image Prior (DIP)~\cite{dip}, which also trains an image-specific network for each test image.\ Averagely, DIP needs $603.9$ seconds to process a $512\times512$ color image, on which our ResNet+NAC network needs $583.2$ seconds (on an NVIDIA Titan X GPU). 

\subsection{Ablation Study}
\label{sec:dis}
To further study our NAC strategy, we conduct more examination of our ResNet+NAC networks on image denoising. Specifically, we assess 1) differences of the ResNet+NAC from the ResNet in DIP~\cite{dip}; 2) how the number of residual blocks and epochs influence the ResNet+NAC; 3) comparison with the ``Oracle'' performance of the ResNet+NAC networks; 4) performance of the ResNet+NAC on ``strong'' noise.



\begin{figure*}[t]
\centering
\subfigure{
\begin{minipage}{0.48\textwidth}
\includegraphics[width=1\textwidth]{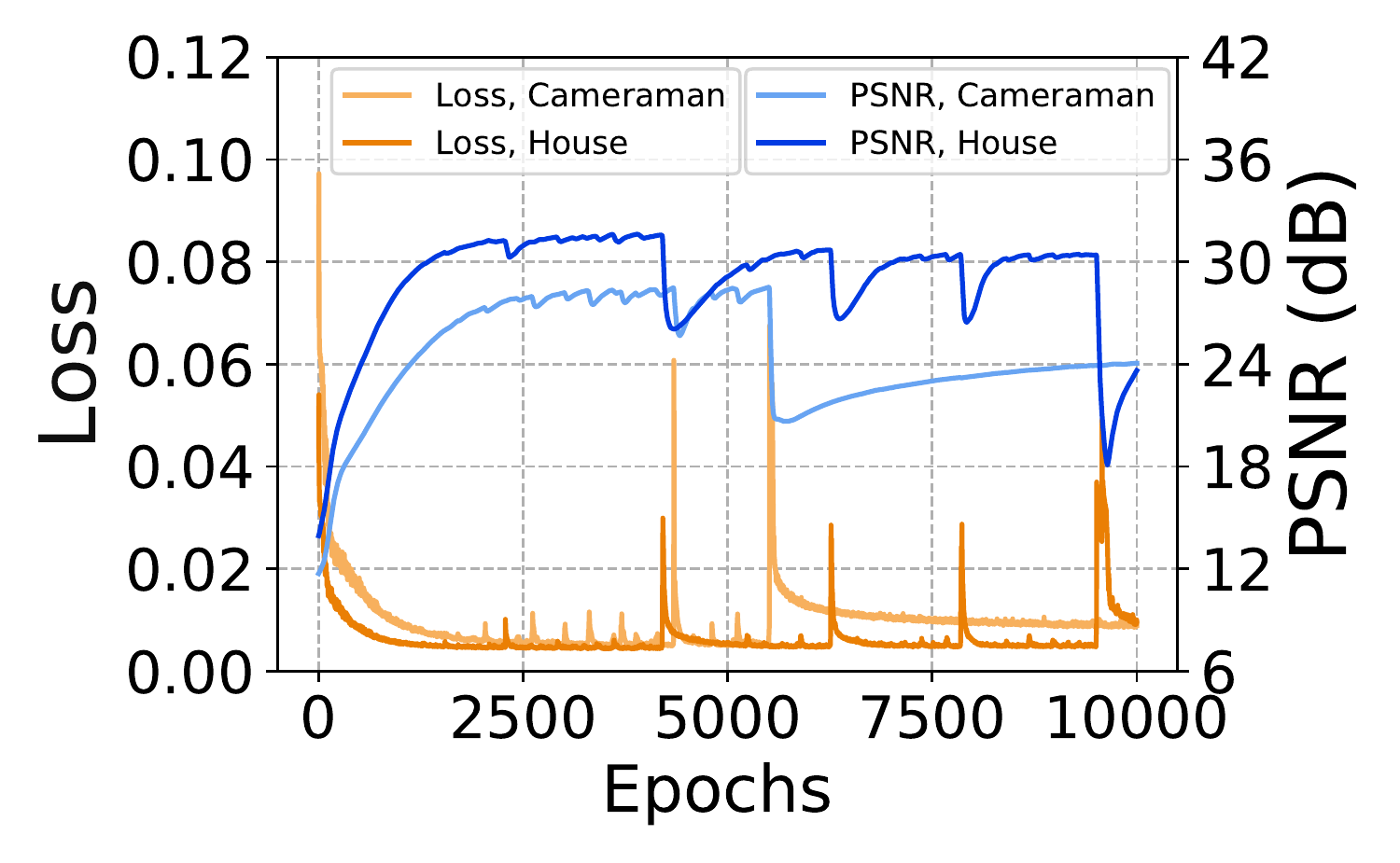}\vspace{-2mm}
\centering{(a) Curves of DIP~\cite{dip}}
\end{minipage}
\begin{minipage}{0.48\textwidth}
\includegraphics[width=1\textwidth]{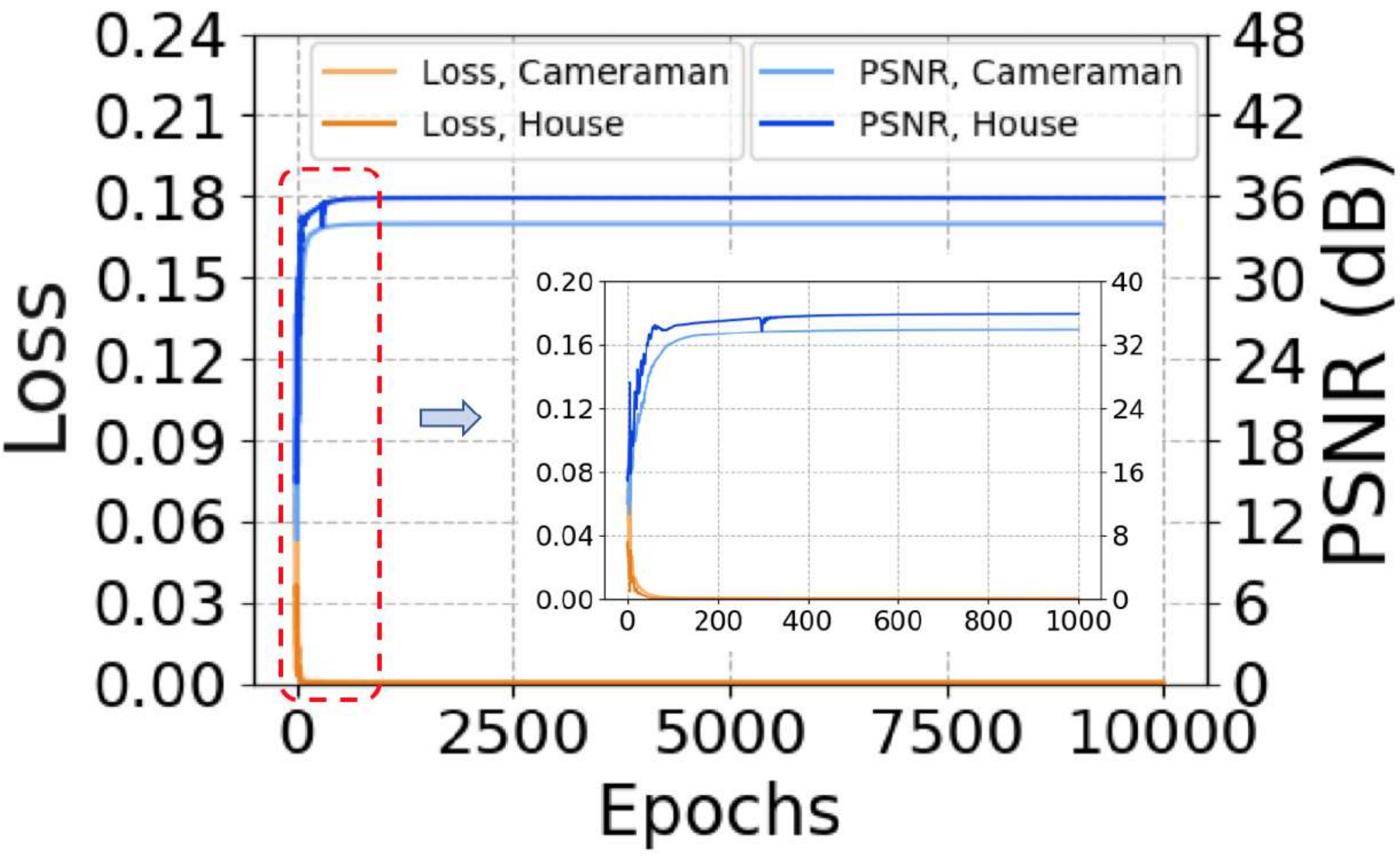}\vspace{-2mm}
\centering{(b) Curves of our NAC}
\end{minipage}
}
\vspace{-2mm}
\caption{\textbf{Training loss and PSNR (dB) curves} of DIP~\cite{dip} (a) and our ResNet+NAC (b) networks w.r.t. the number of epochs, on the images of ``Cameraman'' and ``House'' from \textsl{Set12}.}
\vspace{-2.5mm}
\label{fig:curve}
\end{figure*}

\noindent
\textbf{1) Differences from DIP~\cite{dip}}.\ Though the basic network in our work is the ResNet used in DIP~\cite{dip}, our ResNet+NAC network is essentially different from DIP on at least two aspects.\ First, our ResNet+NAC is a novel strategy for self-supervised learning of \textsl{adaptive network parameters} for the degraded image, while DIP aims to investigate \textsl{adaptive network structure} without learning the parameters.\ Second, our ResNet+NAC learns a mapping from the synthetic noisy image $\mathbf{z}=\mathbf{y}+\mathbf{n}_{s}$ to the noisy image $\mathbf{y}$, which approximates the mapping from the noisy image  $\mathbf{y}=\mathbf{x}+\mathbf{n}_{o}$ to the clean image $\mathbf{x}$.\ But DIP maps a random noise map to the noisy image $\mathbf{y}$, and the denoised image is obtained during the process.\ Due to the two reasons, DIP needs early stop for different images, while our ResNet+NAC achieves more robust (and better) denoising performance than DIP on diverse images.\ In Figure~\ref{fig:curve}, we plot the curves of training loss and test PSNR of DIP (a) and ResNet+NAC (b) networks in 10,000 epochs, on two images of ``Cameraman'' and ``House''.\ We observe that DIP needs early stop to select the best results, while our ResNet+NAC can stably achieve better denoising results within 1000 epochs.
\begin{figure*}[t]
\centering
\subfigure{
\begin{minipage}{0.48\textwidth}
\includegraphics[width=1\textwidth]{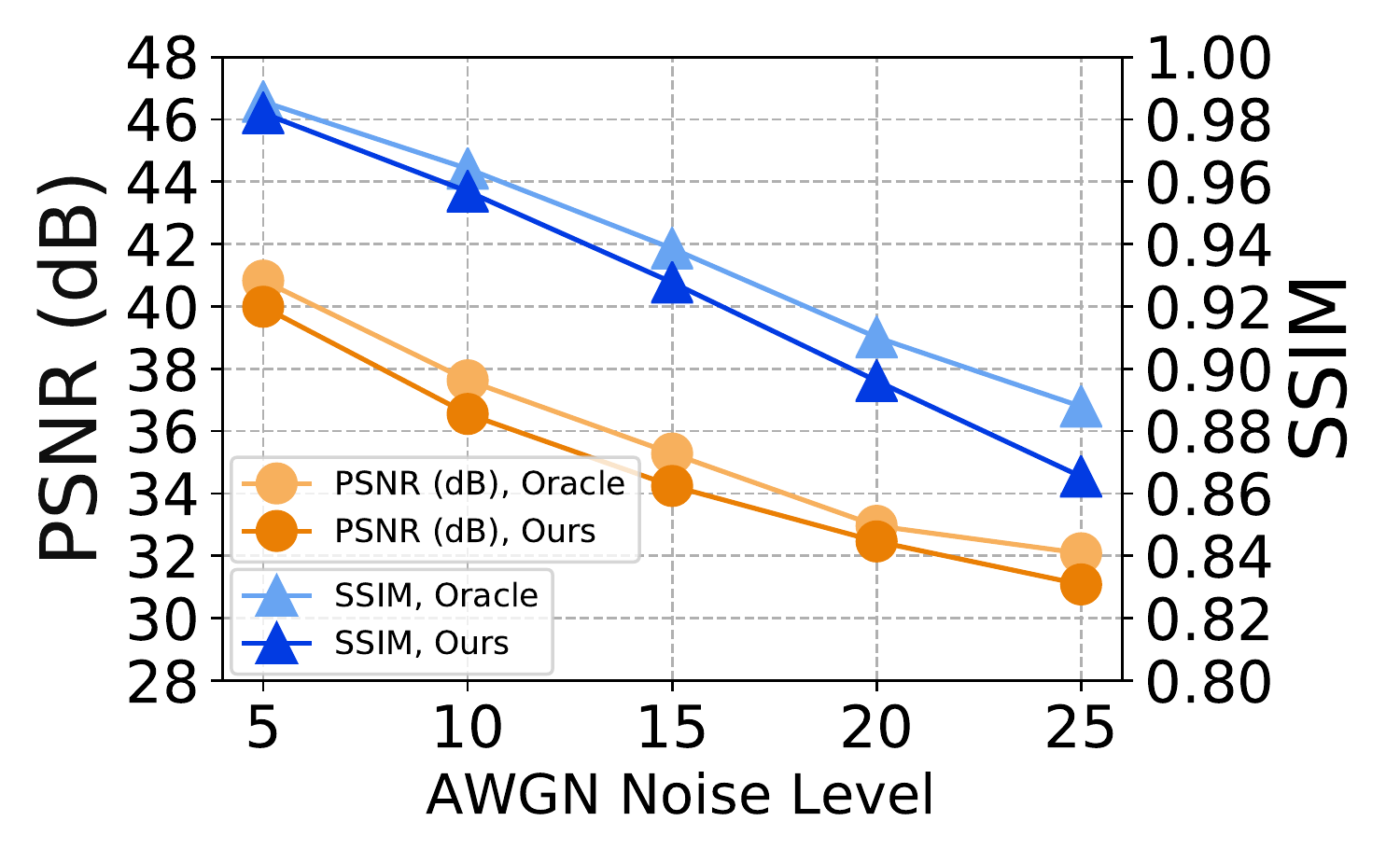}\vspace{-2mm}
\centering{(a)}
\end{minipage}
\begin{minipage}{0.48\textwidth}
\includegraphics[width=1\textwidth]{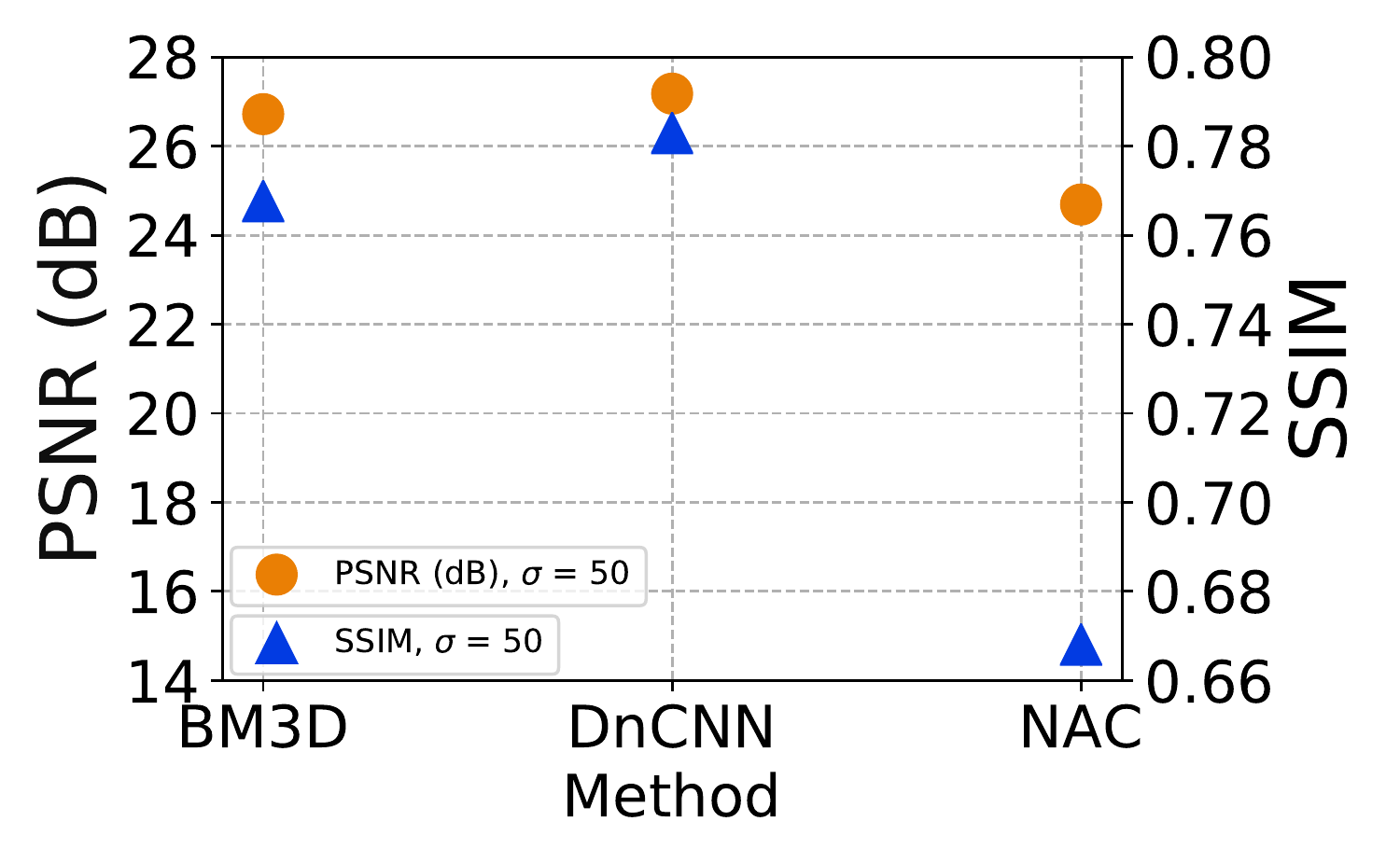}\vspace{-2mm}
\centering{(b)}
\end{minipage}
}
\vspace{-2mm}
\caption{\textbf{Comparisons of PSNR (dB) and SSIM results} on \textsl{Set12} (a) by our ResNet+NAC network and its ``Oracle'' version for AWGN with $\sigma=5,10,15,20,25$ and (b) by BM3D~\cite{bm3d}, DnCNN~\cite{dncnn}, and our ResNet+NAC network for strong AWGN ($\sigma=50$).}
\vspace{-2.5mm}
\label{f-oracle-strong}
\end{figure*}

\noindent
\textbf{2) Influence on the number of residual blocks and epochs}.\ Our backbone network is the ResNet~\cite{dip} with 10 residual blocks trained in 1000 epochs.\ Now we study how the number of residual blocks and epochs influence the performance of ResNet+NAC on image denoising.\ The experiments are performed on the \textsl{Set12} dataset corrupted by AWGN noise ($\sigma=15$).\ From Table~\ref{t-block}, we observe that, with more residual blocks, the ResNet+NAC networks can achieve better PSNR and SSIM~\cite{ssim} results.\ And 10 residual blocks are enough to achieve satisfactory results.\ With more (e.g., 15) blocks, there is little improvement on PSRN and SSIM.\ Hence, we use 10 residual blocks the same as~\cite{dip}.\ Then we study how the number of epochs influence the performance of ResNet+NAC on image denoising.\ From Table~\ref{t-epoch}, one can see that on the \textsl{Set12} dataset corrupted by AWGN noise ($\sigma=15$), with more training epochs, our ResNet+NAC networks achieve better PSNR and SSIM results, but with longer processing time.\
\begin{table}[htp]
\vspace{-0mm}
\centering
\caption{\textbf{Average PSNR (dB)/SSIM of ResNet+NAC with different number of blocks} on \textsl{Set12} corrupted by AWGN noise ($\sigma=15$).}
\renewcommand\arraystretch{1}
\footnotesize
\begin{tabular}{c||cccccc}
\Xhline{1pt}
\rowcolor[rgb]{ .85,  .90,  .95}
\# of Blocks & 1 & 2 & 5 & 10 & 15 
\\
\hline
PSNR$\uparrow$ & 33.58 & 33.85 & 34.14 & 34.24 & 34.26
\\
\hline
SSIM$\uparrow$ & 0.9161 & 0.9226 & 0.9272 & 0.9277 & 0.9272
\\
\hline
\end{tabular}
\vspace{-1mm}
\label{t-block}
\end{table}


\noindent
\textbf{3) Comparison with Oracle}.\ We also study the ``Oracle'' performance of the ResNet+NAC networks.\ In ``Oracle'', we train the ResNet+NAC networks on the pair of \textsl{observed} noisy image $\mathbf{y}$ and its clean image $\mathbf{x}$ corrupted by AWGN noise or signal dependent Poisson noise.\ The experiments are performed on \textsl{Set12} dataset corrupted by AWGN or signal dependent Poisson noise.\ The noise deviations are in $\{5,10,15,20,25\}$.\ Figure~\ref{f-oracle-strong} (a) shows comparisons of our ResNet+NAC and its ``Oracle'' networks on PSNR and SSIM.\ It can be seen that, the ``Oracle'' networks trained on the pair of noisy-clean images only perform slightly better than the original ResNet+NAC networks trained with the \textsl{simulated}-\textsl{observed} noisy image pairs $(\mathbf{z},\mathbf{y})$.\ With our NAC strategy, the ResNet networks trained only with noisy test images achieves similar promising performance on the weak noise.

\noindent
\textbf{4) Performance on strong noise}.\
Our NAC strategy is based on the assumption of ``weak noise''.\ It is natural to wonder how well ResNet+NAC performs against strong noise.\ To answer this question, we compare the ResNet+NAC networks with BM3D~\cite{bm3d} and DnCNN~\cite{dncnn}, on \textsl{Set12} corrupted by AWGN noise with $\sigma=50$.\ The PSNR and SSIM results are plotted in Figure~\ref{f-oracle-strong} (b).\ One can see that, our ResNet+NAC networks are limited in handling strong AWGN noise, when compared with BM3D~\cite{bm3d} and DnCNN~\cite{dncnn}.
\begin{table}[htp]
\vspace{-0mm}
\centering
\caption{\textbf{Average PSNR (dB) and time (s) of ResNet+NAC with different number of epochs} on \textsl{Set12} corrupted by AWGN noise ($\sigma=15$).}
\renewcommand\arraystretch{1}
\footnotesize
\begin{tabular}{c||ccccc}
\Xhline{1pt}
\rowcolor[rgb]{ .85,  .90,  .95}
\# of Epochs & 100 & 200 & 500 & 1000 & 5000
\\
\hline
PSNR$\uparrow$ & 31.80 & 32.79 & 33.77 & 34.24 & 34.28 
\\
\hline
SSIM$\uparrow$ & 0.8714 & 0.9023 & 0.9189 & 0.9277 & 0.9280
\\
\hline
Time$\downarrow$ & 67.4& 132.5 & 302.0 & 583.2 & 2815.6
\\
\hline
\end{tabular}
\vspace{-1mm}
\label{t-epoch}
\end{table}

\section{Conclusion}
\label{sec:con}
In this work, we proposed a ``Noisy-As-Clean'' (NAC) strategy for learning self-supervised image denoising networks.\ In our NAC, we trained an image-specific network by taking the corrupted image as the target, and adding to it the simulated noise to generate the doubly corrupted noisy input.\ The simulated noise is close to the observed noise in the noisy test image.\ This strategy can be seamlessly embedded into existing supervised denoising networks.\ We observed that \textsl{it is possible to learn a self-supervised network only with the corrupted image, approximating the optimal parameters of a supervised network learned with a pair of noisy and clean images}.\ Extensive experiments on synthetic and real-world benchmarks demonstrate that, the DnCNN~\cite{dncnn} and ResNet in Deep Image Prior~\cite{dip} trained with our NAC strategy achieved comparable or better performance on PSNR, SSIM, and visual quality, when compared to previous state-of-the-art image denoising methods, including supervised denoising networks.\ These results validate that our NAC strategy can learn effctive image-specific priors and noise statistics only from the corrupted test image.





{
\small\small
\bibliographystyle{ieee} 
\bibliography{NAC}

\begin{thebibliography}{10}\itemsep=-1pt

\bibitem{sidd2018}
A.~Abdelhamed, S.~Lin, and M.~S. Brown.
\newblock A high-quality denoising dataset for smartphone cameras.
\newblock In {\em CVPR}, June 2018.

\bibitem{neatimage}
N.~ABSoft.
\newblock Neat {I}mage.
\newblock \url{https://ni.neatvideo.com/home}.

\bibitem{noise2self}
J.~Batson and L.~Royer.
\newblock {N}oise2{S}elf: Blind denoising by self-supervision.
\newblock In {\em ICML}, volume~97, pages 524--533. PMLR, 2019.

\bibitem{billingsley1995probability}
P.~Billingsley.
\newblock {\em Probability and Measure}.
\newblock Wiley Series in Probability and Statistics. Wiley, 1995.

\bibitem{upi}
T.~Brooks, B.~Mildenhall, T.~Xue, J.~Chen, D.~Sharlet, and J.~T. Barron.
\newblock Unprocessing images for learned raw denoising.
\newblock In {\em CVPR}, pages 9446--9454, 2019.

\bibitem{mlp}
H.~C. Burger, C.~J. Schuler, and S.~Harmeling.
\newblock Image denoising: Can plain neural networks compete with {BM3D}?
\newblock In {\em CVPR}, pages 2392--2399, 2012.

\bibitem{gcbd}
J.~Chen, J.~Chen, H.~Chao, and M.~Yang.
\newblock Image blind denoising with generative adversarial network based noise
  modeling.
\newblock In {\em CVPR}, pages 3155--3164, 2018.

\bibitem{chen2017trainable}
Y.~Chen and T.~Pock.
\newblock Trainable nonlinear reaction diffusion: A flexible framework for fast
  and effective image restoration.
\newblock {\em IEEE Transactions on Pattern Analysis and Machine Intelligence},
  39(6):1256--1272, 2017.

\bibitem{cbm3d}
K.~Dabov, A.~Foi, V.~Katkovnik, and K.~Egiazarian.
\newblock Color image denoising via sparse 3{D} collaborative filtering with
  grouping constraint in luminance-chrominance space.
\newblock In {\em ICIP}, pages 313--316. IEEE, 2007.

\bibitem{bm3d}
K.~Dabov, A.~Foi, V.~Katkovnik, and K.~Egiazarian.
\newblock Image denoising by sparse 3-{D} transform-domain collaborative
  filtering.
\newblock {\em IEEE Transactions on Image Processing}, 16(8):2080--2095, 2007.

\bibitem{cvid2020}
Y.~Du, J.~Xu, X.~Zhen, M.-M. Cheng, and L.~Shao.
\newblock Conditional variational image deraining.
\newblock {\em IEEE Transactions on Image Processing}, pages 6288--6301, 2020.

\bibitem{ksvd}
M.~Elad and M.~Aharon.
\newblock Image denoising via sparse and redundant representations over learned
  dictionaries.
\newblock {\em IEEE Transactions on Image Processing}, 15(12):3736--3745, 2006.

\bibitem{poissongaussian}
A.~Foi, M.~Trimeche, V.~Katkovnik, and K.~Egiazarian.
\newblock Practical poissonian-gaussian noise modeling and fitting for
  single-image raw-data.
\newblock {\em IEEE Transactions on Image Processing}, 17(10):1737--1754, Oct
  2008.

\bibitem{cbdnet}
S.~Guo, Z.~Yan, K.~Zhang, W.~Zuo, and L.~Zhang.
\newblock Toward convolutional blind denoising of real photographs.
\newblock In {\em CVPR}, 2019.

\bibitem{resnet}
K.~He, X.~Zhang, S.~Ren, and J.~Sun.
\newblock Deep residual learning for image recognition.
\newblock In {\em CVPR}, pages 770--778, 2016.

\bibitem{NLH2020}
Y.~Hou, J.~Xu, M.~Liu, G.~Liu, L.~Liu, F.~Zhu, and L.~Shao.
\newblock Nlh: A blind pixel-level non-local method for real-world image
  denoising.
\newblock {\em IEEE Transactions on Image Processing}, 20(1):5121--5135, 2020.

\bibitem{bn2015}
S.~Ioffe and C.~Szegedy.
\newblock Batch normalization: Accelerating deep network training by reducing
  internal covariate shift.
\newblock In {\em ICML}, 2015.

\bibitem{adam}
D.~P. Kingma and J.~Ba.
\newblock Adam: A method for stochastic optimization.
\newblock In {\em ICLR}, 2015.

\bibitem{noise2void}
A.~Krull, T.-O. Buchholz, and F.~Jug.
\newblock Noise2{V}oid-learning denoising from single noisy images.
\newblock In {\em CVPR}, 2019.

\bibitem{ss2019}
S.~Laine, T.~Karras, J.~Lehtinen, and T.~Aila.
\newblock High-quality self-supervised deep image denoising.
\newblock In {\em NeurIPS}, 2019.

\bibitem{nlnet}
S.~Lefkimmiatis.
\newblock Non-local color image denoising with convolutional neural networks.
\newblock In {\em CVPR}, pages 3587--3596, 2017.

\bibitem{noise2noise}
J.~Lehtinen, J.~Munkberg, J.~Hasselgren, S.~Laine, T.~Karras, M.~Aittala, and
  T.~Aila.
\newblock Noise2{N}oise: Learning image restoration without clean data.
\newblock In {\em ICML}, pages 2971--2980, 2018.

\bibitem{dip}
V.~Lempitsky, D.~U. Andrea~Vedaldi, and V.~Lempitsky.
\newblock Deep image prior.
\newblock In {\em CVPR}, pages 9446--9454, 2018.

\bibitem{Liang_2018_CVPR}
Z.~Liang, J.~Xu, D.~Zhang, Z.~Cao, and L.~Zhang.
\newblock A hybrid l1-l0 layer decomposition model for tone mapping.
\newblock In {\em The IEEE Conference on Computer Vision and Pattern
  Recognition (CVPR)}, June 2018.

\bibitem{liu2006noise}
C.~Liu, W.~T. Freeman, R.~Szeliski, and S.~B. Kang.
\newblock Noise estimation from a single image.
\newblock {\em CVPR}, 1:901--908, 2006.

\bibitem{nlrn2018}
D.~Liu, B.~Wen, Y.~Fan, C.~C. Loy, and T.~S. Huang.
\newblock Non-local recurrent network for image restoration.
\newblock In {\em NeurIPS}, pages 1673--1682, 2018.

\bibitem{relu2010}
V.~Nair and G.~E. Hinton.
\newblock Rectified linear units improve restricted boltzmann machines.
\newblock In {\em ICML}, pages 807--814, 2010.

\bibitem{crosschannel2016}
S.~Nam, Y.~Hwang, Y.~Matsushita, and S.~J. Kim.
\newblock A holistic approach to cross-channel image noise modeling and its
  application to image denoising.
\newblock {\em In CVPR}, pages 1683--1691, 2016.

\bibitem{dnd2017}
T.~Pl{\"o}tz and S.~Roth.
\newblock Benchmarking denoising algorithms with real photographs.
\newblock In {\em CVPR}, 2017.

\bibitem{n3net}
T.~Pl{\"o}tz and S.~Roth.
\newblock Neural nearest neighbors networks.
\newblock In {\em NeurIPS}, 2018.

\bibitem{foe}
S.~Roth and M.~J. Black.
\newblock Fields of experts.
\newblock {\em International Journal of Computer Vision}, 82(2):205--229, 2009.

\bibitem{vggnet}
K.~Simonyan and A.~Zisserman.
\newblock Very deep convolutional networks for large-scale image recognition.
\newblock In {\em ICLR}, 2015.

\bibitem{googlenet}
C.~Szegedy, W.~Liu, Y.~Jia, P.~Sermanet, S.~Reed, D.~Anguelov, D.~Erhan,
  V.~Vanhoucke, and A.~Rabinovich.
\newblock Going deeper with convolutions.
\newblock In {\em CVPR}, pages 1--9, 2015.

\bibitem{ssim}
Z.~Wang, A.~C. Bovik, H.~R. Sheikh, and E.~P. Simoncelli.
\newblock Image quality assessment: from error visibility to structural
  similarity.
\newblock {\em IEEE Transactions on Image Processing}, 13(4):600--612, 2004.

\bibitem{STAR2020}
J.~Xu, Y.~Hou, D.~Ren, L.~Liu, F.~Zhu, M.~Yu, H.~Wang, and L.~Shao.
\newblock Star: A structure and texture aware retinex model.
\newblock {\em IEEE Transactions on Image Processing}, 29:5022--5037, 2020.

\bibitem{PolyUdataset}
J.~Xu, H.~Li, Z.~Liang, D.~Zhang, and L.~Zhang.
\newblock Real-world noisy image denoising: {A} new benchmark.
\newblock {\em CoRR}, abs/1804.02603, 2018.

\bibitem{gid2018}
J.~Xu, L.~Zhang, and D.~Zhang.
\newblock External prior guided internal prior learning for real-world noisy
  image denoising.
\newblock {\em IEEE Transactions on Image Processing}, 27(6):2996--3010, June
  2018.

\bibitem{twsc}
J.~Xu, L.~Zhang, and D.~Zhang.
\newblock A trilateral weighted sparse coding scheme for real-world image
  denoising.
\newblock In {\em ECCV}, 2018.

\bibitem{mcwnnm}
J.~Xu, L.~Zhang, D.~Zhang, and X.~Feng.
\newblock Multi-channel weighted nuclear norm minimization for real color image
  denoising.
\newblock In {\em ICCV}, 2017.

\bibitem{pgpd}
J.~Xu, W.~Zuo, L.~Zhang, D.~Zhang, and X.~Feng.
\newblock Patch group based nonlocal self-similarity prior learning for image
  denoising.
\newblock In {\em ICCV}, pages 244--252, 2015.

\bibitem{dncnn}
K.~Zhang, W.~Zuo, Y.~Chen, D.~Meng, and L.~Zhang.
\newblock Beyond a {Gaussian} denoiser: Residual learning of deep {CNN} for
  image denoising.
\newblock {\em IEEE Transactions on Image Processing}, 26(7):3142--3155, 2017.

\bibitem{iraniinternal}
M.~Zontak and M.~Irani.
\newblock Internal statistics of a single natural image.
\newblock In {\em CVPR}, 2011.

\bibitem{epll}
D.~Zoran and Y.~Weiss.
\newblock From learning models of natural image patches to whole image
  restoration.
\newblock In {\em ICCV}, pages 479--486, 2011.

\end{thebibliography}
}

\end{document}